\documentclass{article}

\newif\ifappendix
\appendixtrue % Comment to remove appendix

\usepackage{amsmath,amsthm,amssymb}
\usepackage{mathtools}
\usepackage{bm}
\usepackage{dsfont}
\usepackage{xspace}
\usepackage{color,soul}
\usepackage{multirow}
\usepackage{multicol}
\usepackage{caption}
\usepackage{enumitem}
\usepackage{makecell}

\newcommand{\sys}{NAFS\xspace}

\newcommand{\para}[1]{{\vspace{2pt} \bf \noindent #1 \hspace{1pt}}}
\usepackage{CJKutf8}

\newtheorem{theorem}{Theorem}

\newtheorem{lemma}[theorem]{Lemma}

\newtheorem{definition}{Definition}

\newtheorem*{assumption*}{\assumptionnumber}
\providecommand{\assumptionnumber}{}


\newtheorem*{remark*}{\remarknumber}
\providecommand{\remarknumber}{}


\usepackage{microtype}
\usepackage{graphicx}
\usepackage{subfigure}
\usepackage{booktabs} % for professional tables

\usepackage{hyperref}

\usepackage[linesnumbered,vlined,ruled,noend]{algorithm2e}
\usepackage[accepted]{icml2022}

\icmltitlerunning{\sys: A Simple yet Tough-to-beat Baseline for Graph Representation Learning}
\ifappendix
\icmltitlerunning{\sys: A Simple yet Tough-to-beat Baseline for Graph Representation Learning (with appendix)}
\else
\icmltitlerunning{\sys: A Simple yet Tough-to-beat Baseline for Graph Representation Learning}
\fi

\begin{document}

\twocolumn[
\icmltitle{\sys: A Simple yet Tough-to-beat Baseline for Graph Representation Learning}
% \ifappendix
% \icmltitle{Don't Waste Your Bits! Squeeze Activations and Gradients for\\Deep Neural Networks via \alg (with appendix)}
% \else
% \icmltitle{Don't Waste Your Bits! Squeeze Activations and Gradients for\\Deep Neural Networks via \alg}
% \fi



\icmlsetsymbol{equal}{*}

\begin{icmlauthorlist}
\icmlauthor{Wentao Zhang}{pku}
\icmlauthor{Zeang Sheng}{pku}
\icmlauthor{Mingyu Yang}{pku}
\icmlauthor{Yang Li}{pku}
\icmlauthor{Yu Shen}{pku}
\icmlauthor{Zhi Yang}{pku}
\icmlauthor{Bin Cui}{pku,qd}
\end{icmlauthorlist}

\icmlaffiliation{pku}{School of CS \& Key Laboratory of High Confidence Software Technologies, Peking University}
% \icmlaffiliation{pku}{Peking University}
\icmlaffiliation{qd}{Institute of Computational Social Science, Peking University (Qingdao), China}
% \icmlaffiliation{bupt}{Beijing University of Posts and Telecommunications}
% \icmlaffiliation{bupt}{Beijing Key Lab of Intelligent Telecommunications Software and Multimedia, BUPT}
% \icmlaffiliation{tencent}{Tencent Inc.}
% \icmlaffiliation{pku_bigdata}{Center for Data Science \& National Engineering Laboratory for Big Data Analysis and Applications, Peking University}

\icmlcorrespondingauthor{Bin Cui}{bin.cui@pku.edu.cn}
\icmlkeywords{Machine Learning, ICML}

\vskip 0.3in
]



\printAffiliationsAndNotice{}  % leave blank if no need to mention equal contribution

% Recently, graph neural networks (GNNs) have shown prominent performance in graph representation learning by leveraging knowledge from both graph structure and node features. However, most of them have two major limitations. First, GNNs can learn higher-order structural information by stacking more layers but can not deal with large depth due to the over-smoothing issue. Second, it is not easy to apply these methods on large graphs due to the expensive computation cost and high memory usage.
% In this paper, we present node-adaptive feature smoothing (NAFS), a simple non-parametric method that constructs node representations without parameter learning. NAFS first extracts the features of each node with its neighbors of different hops by feature smoothing, and then adaptively combines the smoothed features. Besides, the constructed node representation can further be enhanced by the ensemble of smoothed features extracted via different smoothing strategies. We conduct experiments on four benchmark datasets on two different application scenarios: node clustering and link prediction.
% Remarkably, NAFS with feature ensemble outperforms the state-of-the-art GNNs on these tasks and mitigates the aforementioned two limitations of most learning-based GNN counterparts. 

\begin{abstract}
Recently, graph neural networks (GNNs) have shown prominent performance in graph representation learning by leveraging knowledge from both graph structure and node features.
However, most of them have two major limitations.
First, GNNs can learn higher-order structural information by stacking more layers but can not deal with large depth due to the over-smoothing issue. 
Second, it is not easy to apply these methods on large graphs due to the expensive computation cost and high memory usage.
In this paper, 
we present node-adaptive feature smoothing (NAFS), a simple non-parametric method that constructs node representations {\em without parameter learning}.
\sys first extracts the features of each node with its neighbors of different hops by {\em feature smoothing}, and then adaptively combines the smoothed features. 
Besides, the constructed node representation can further be enhanced by the ensemble of smoothed features extracted via different smoothing strategies.
We conduct experiments on four benchmark datasets on two different application scenarios: node clustering and link prediction.
Remarkably, \sys with feature ensemble outperforms the state-of-the-art GNNs on these tasks and mitigates the aforementioned two limitations of most learning-based GNN counterparts. 
\end{abstract}

% Recently, graph neural networks (GNNs) have shown prominent performance in graph representation learning by leveraging knowledge from both graph structure and node features. However, most of them have two major limitations. First, GNNs can learn higher-order structural information by stacking more layers but can not deal with large depth due to the over-smoothing issue. Second, it is not easy to apply these methods on large graphs due to the expensive computation cost and high memory usage. In this paper, we present node-adaptive feature smoothing (NAFS), a simple non-parametric method that constructs node representations without parameter learning. NAFS first extracts the features of each node with its neighbors of different hops by feature smoothing, and then adaptively combines the smoothed features. Besides, the constructed node representation can further be enhanced by the ensemble of smoothed features extracted via different smoothing strategies. We conduct experiments on four benchmark datasets on two different application scenarios: node clustering and link prediction. Remarkably, NAFS with feature ensemble outperforms the state-of-the-art GNNs on these tasks and mitigates the aforementioned two limitations of most learning-based GNN counterparts. 

\section{Introduction}
In recent years, graph representation learning has been extensively applied in various application scenarios, such as node clustering, link prediction, node classification, and graph classification~\citep{kipf2016variational, wu2020graph, zhang2020reliable, miao2021degnn, jiang2022zoomer, wang2016structural, wu2020garg, miao2021lasagne}. 
The goal of graph representation learning is to encode graph information to node embeddings.
Traditional graph representation learning methods, such as DeepWalk~\citep{perozzi2014deepwalk}, Node2vec~\citep{grover2016node2vec}, LINE ~\citep{tang2015line}, and ComE~\citep{cavallari2017learning}
merely focus on preserving graph structure information.  
GNN-based graph representation learning has
attracted intensive interest by combining knowledge from both graph structure and node features.
While most of these GNN-based methods are designed based on Graph AutoEncoder (GAE) and Variational Graph AutoEncoder (VGAE)~\citep{kipf2016variational}, these methods share two major limitations:
%, which have been successfully applied to several challenging graph representation learning tasks.

\textbf{Shallow Architecture.} 
Previous work shows that although stacking multiple GNN layers in Graph Convolutional Network (GCN)~\citep{kipf2016semi} is capable of exploiting deep structural information, applying a large number of GNN layers might lead to indistinguishable node embeddings, i.e., the over-smoothing issue~\citep{li2018deeper, zhang2021evaluating}.
% the feature aggregation operation in Graph Convolutional Network (GCN)~\citep{DBLP:conf/iclr/KipfW17} is exactly Laplacian smoothing, and stacking $k$ GCN layers enables the model to acquire $k$-hop neighborhood information.
%However, large $k$ will introduce the over-smoothing issue~\citep{li2018deeper}, which leads to indistinguishable node embeddings.
Therefore, most state-of-the-art GNNs resort to shallow architectures, which hinders the model from capturing long-range dependencies.
% \red{In addition, many GNN-based graph representation learning methods (e.g., GAE) forces each node to recover only its directed neighbors, which might ignore the high order neighborhood information. --sy: What is the meaning of ``recover'' here? And, this paragraph is about shallow architecture.}

\textbf{Low Scalability.} GNN-based graph representation learning methods can not scale well to large graphs due to the expensive computation cost and high memory usage.
%\sza{后面没强调，对应的是scalability的实验，abstract里的说明同理}
% Most existing methods need to repeatedly perform recursive feature smoothing at each training epoch with the participation of 
% % all the nodes and the whole adjacency matrix.
% the entire graph.
Most existing GNNs need to repeatedly perform the computationally expensive and recursive feature smoothing, which involves the participation of 
% all the nodes and the whole adjacency matrix.
the entire graph at each training epoch.~\citep{zhang2022pasca}
Furthermore, most methods adopt the same training loss function as GAE, which introduces high memory usage by storing the dense-form adjacency matrix on GPU.
For a graph of size 200 million, its dense-form adjacency matrix requires a space of roughly 150GB, exceeding the memory capacity of the current powerful GPU devices.
%This process is hard to scale up due to the expensive computation and communication cost, especially for large-scale graphs in distributed environments. 
%For example, the current largest public graph data -- ogbn-papers100M~\citep{hu2020ogb} consists of 111 million nodes and more than 1.6 billion edges. 
%It is especially hard to train GNNs on such a large graph due to high memory usage and computation cost.

To tackle these issues, we propose a new graph representation learning method, which is embarrassingly simple: just smooth the node features and then combine the smoothed features in a node-adaptive manner.
We name this method node-adaptive feature smoothing (NAFS), and its goal is to construct better node embeddings that integrate the information from both graph structural information and node features.
Based on the observation that different nodes have highly diverse ``smoothing speed'', \sys adaptively smooths each node feature and takes advantage of both low-order and high-order neighborhood information of each node.
In addition, feature ensemble is also employed to combine the smoothed features extracted via different smoothing operators.
Since \sys is training-free, it significantly reduces the training cost and scales better to large graphs than most GNN-based graph representation learning methods.

This paper is not meant to diminish the current advancements in GNN-based graph representation learning approaches.
Instead, we aim to introduce an easier way to obtain high-quality node embeddings and understand the source of performance gains of these approaches better. 
Feature smoothing could be a promising direction towards a more simple and effective integration of information from both graph structure and node features.

Our contributions are as follows: 
(1) \textit{\underline{New perspective}}. 
% To the best of our knowledge, we are the first to present the incredible results that simple feature smoothing without any trainable parameters could even outperform state-of-the-art GNNs, which points out a new direction towards efficient and scalable graph representation learning.
To the best of our knowledge, we are the first to explore the possibility that simple feature smoothing without any trainable parameters could even outperform state-of-the-art GNNs; this incredible finding opens up a new direction towards efficient and scalable graph representation learning. 
(2) \textit{\underline{Novel method}}. 
We propose \sys, a node-adaptive feature smoothing approach along with various feature ensemble strategies, to fully exploit knowledge from both the graph structure and node features.
%Our ablation study proves that both these two methods are effective to capture more powerful node features.
%(3) \textit{\underline{Theoretical analysis}}.  We make a detailed analysis about the mechanism of feature smoothing and display two factors affecting our result of \sys: the sparsity of graph and the node degree. 
(3) \textit{\underline{State-of-the-art performance}}. 
We evaluate the {\em effectiveness} and {\em efficiency} of \sys on different datasets and graph-based tasks, including node clustering and link prediction. 
Empirical results demonstrate that \sys performs comparably with or even outperforms the state-of-the-art GNNs, and achieves up to two orders of magnitude speedup.
In particular, on PubMed, \sys outperforms GAE~\citep{kipf2016variational} and AGE~\citep{cui2020adaptive} by a margin of $9.0\%$ and $3.8\%$ in terms of NMI in node clustering, while achieving up to $65.4\times$ and $88.6\times$ training speedups, respectively.

\section{Preliminary}
In this section, we first explain the notations and problem formulation. Then, we review current GNNs and GNN-based graph representation learning.

\subsection{Notations and Problem Formulation.}
In this paper, we consider an undirected graph $\mathcal{G}$ = ($\mathcal{V}$,$\mathcal{E}$) with $|\mathcal{V}| = n$ nodes and $|\mathcal{E}| = m$ edges. 
Here we suppose that $m \propto n$ as it is the case in most real-world graphs. 
We denote by $\mathbf{A}$ the adjacency matrix of $\mathcal{G}$. Each node can possibly have a feature vector of size $f$, which stacks up to an $n \times f$ feature matrix $\mathbf{X}$. 
The degree matrix of $\mathbf{A}$  is denoted as $\mathbf{D}=\operatorname{diag}\left(d_{1}, d_{2}, \cdots, d_{n}\right) \in \mathbb{R}^{n \times n}$, where $d_{i}=\sum_{v_{j} \in \mathcal{V}} \mathbf{A}_{i j}$. 
We denote the final node embedding matrix as $\mathbf{Z}$, and evaluate it in both the node clustering and the link prediction tasks.
%We denote $\mathbf{Z}$ as the embedding matrix, and the embeddings preserve both the structure and feature information of graph $\mathcal{G}$. 
The node clustering task requires the model to partition the nodes into $c$ disjoint groups ${G_{1},G_{2},\cdots,G_{c}}$, where similar nodes should be in the same group. 
The target of the link prediction task is to predict whether an edge exists between given node pairs.

% \subsection{Feature Engineering}
% \zwt{说一下我们和传统特征工程的区别，图特征工程目标：旨在通过一种training-free的方式来高效融合feature和邻接矩阵adj，从而产生更powerful的新feature}

%\sza{可能要说一下GNN和GCN的关系，GNN用形式化的方式表示，应该要用graphsage或者mpnn里的表示方法，这篇paper里用的形式化的表示方法比较局限，后续说明的时候比较别扭}
\subsection{Graph Convolutional Network.}
% \subsubsection{Semi-GNNs}
% \zwt{给GNN做分类，然后diss scalability和depth等问题}
Based on the assumption that locally connected nodes are likely to enjoy high similarity~\citep{mcpherson2001birds}, each node in most GNN models iteratively smooths the representations of its neighbors for better node embedding.
Below is the formula of the $l$-th GCN layer~\citep{kipf2016semi}:
%The commonly used GNN layer updates the node feature embedding in the graph by aggregating the features of neighboring nodes:
\begin{small}
\begin{equation}
    \small
    \mathbf{X}^{(l)}=\delta\big(\mathbf{\hat{A}}\mathbf{X}^{(l-1)}\mathbf{\Theta}^{(l)}\big), \quad
    \mathbf{\hat{A}} = \widetilde{\mathbf{D}}^{-1/2}\widetilde{\mathbf{A}}\widetilde{\mathbf{D}}^{-1/2}, 
    \label{eq_GC}
\end{equation}
\end{small}
where $\mathbf{\widetilde{A}}=\mathbf{A}+\mathbf{I}_n$ is the adjacency matrix of the undirected graph $\mathcal{G}$ with self loop added, $\mathbf{X}^{(l)}$ is the node embedding matrix at layer $l$, $\mathbf{X}^{(0)}$ is the original feature matrix, $\mathbf{\Theta}^{(l)}$ are the trainable weights, and $\delta$ is the activation function.
%$\widetilde{\mathbf{A}}$ is the adjacency matrix of the undirected graph $\mathcal{G}$ with self loop added.
$\mathbf{\hat{A}}$ is the smoothing matrix that helps each node to smooth representations of neighboring nodes.
% By setting $r = $ 0.5, 1 and 0, $\mathbf{\hat{A}}=\widetilde{\mathbf{D}}^{r-1}\widetilde{\mathbf{A}}\widetilde{\mathbf{D}}^{-r}$ represents the symmetric normalized adjacency matrix $\widetilde{\mathbf{D}}^{-1/2}\widetilde{\mathbf{A}}\widetilde{\mathbf{D}}^{-1/2}$~\citep{DBLP:conf/iclr/KlicperaBG19}, the random walk transition probability matrix $\widetilde{\mathbf{D}}^{-1}\widetilde{\mathbf{A}}$~\citep{xu2018representation}, and the reverse transition probability matrix $\widetilde{\mathbf{A}}\widetilde{\mathbf{D}}^{-1}$~\citep{DBLP:conf/iclr/ZengZSKP20}, respectively. 
% By setting different parameter $r$ in $\widetilde{\mathbf{D}}^{r-1}\widetilde{\mathbf{A}}\widetilde{\mathbf{D}}^{-r}$, different strategies can be adopted to aggregate the neighborhood information.

\begin{figure*}
	\centering
	\includegraphics[width=.7\linewidth]{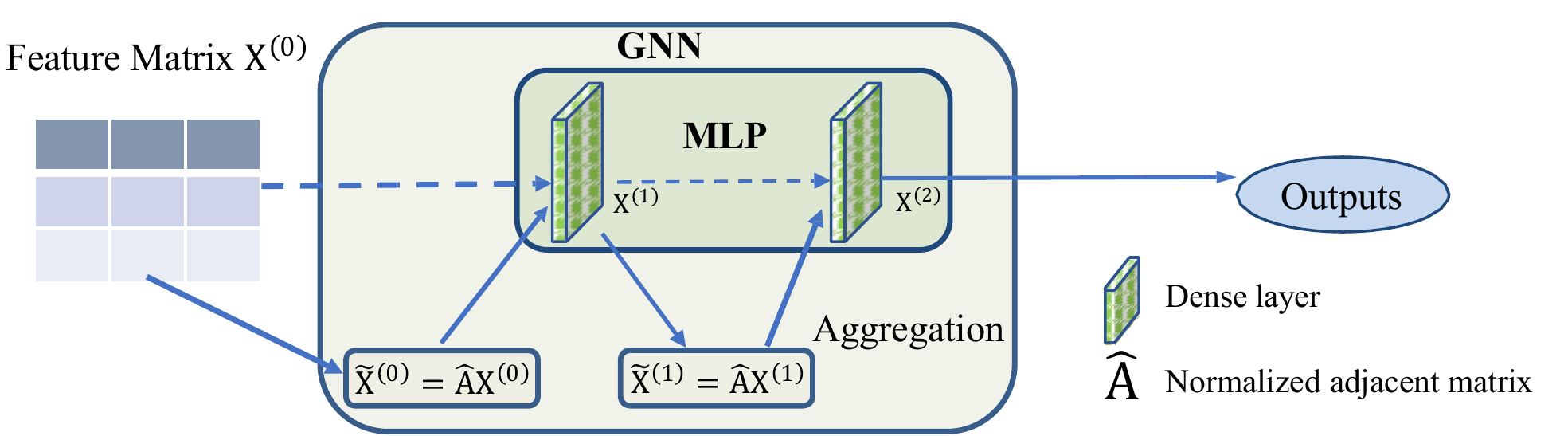}
	\caption{\small  The framework of a two-layer GCN models.}
	\label{fig:GCN}
% 	\vspace{-5mm}
\end{figure*}

As shown in Eq.~\ref{eq_GC}, each GCN layer contains two operations: feature aggregation (smoothing) and feature transformation.
Figure~\ref{fig:GCN} shows the framework of a two-layer GCN.
The $l$-th layer in GCN firstly executes feature smoothing on the node embedding $\mathbf{X}^{(l-1)}$. 
Then, the smoothed feature $\mathbf{\widetilde{X}}^{(l-1)}$ is transformed with trainable weights $\mathbf{\Theta}^{(l)}$ and activation function $\delta$ to generate new node embedding $\mathbf{X}^{(l)}$.
Note that GCN will degrade to MLP if feature smoothing is removed from each layer.

\subsection{GNN-based Graph Representation Learning.}
\label{GAE_related}
% \subsection{Co}
%\zwt{介绍GAE，摆公式}
%\zwt{主要就是做graph embedding，讲一下传统的embedding和GNN-based Embedding-参考AGE，以及它在link prediction和聚类的应用}
%\zwt{说我们和Graph Embedding方法的主要区别就是embedding一般都是训练得到的并且常常伴随着降维，而我们无需训练也不改变feature维度}
GAE~\citep{kipf2016variational}, the first and the most representative GNN-based graph embedding method, adopts an encoder to generate node embedding matrix $\mathbf{Z}$ with inputs $\mathbf{\hat{A}}$ and $\mathbf{X}$.
A simple inner product decoder is then used to reconstruct the adjacency matrix.
The final training loss of GAE is the binary entropy loss between $\mathbf{A}'$ and $\mathbf{\widetilde{A}}$, the reconstructed adjacency matrix and the original adjacency matrix with self loop added. Specifically, the loss function can be defined as follow
\begin{small}
\begin{equation}
    \mathcal{L}=\sum_{1 \leq i, j \leq n}-\mathbf{\widetilde{A}}_{i,j}\log \mathbf{A}'_{i,j}-(1-\mathbf{\widetilde{A}}_{i,j}) \log(1-\mathbf{A}'_{i,j})),
\end{equation}
\label{reconstruct}
\end{small}
%\sza{下面这句话变成公式，单独一句话太难看}
%The final training loss function of GAE is the binary cross entropy loss between $\mathbf{A}$ and $\mathbf{A}'$.
\noindent Where $\mathbf{A}'=\text{sigmoid}(\mathbf{Z}\cdot\mathbf{Z}^\mathsf{T})$ is the reconstructed adjacency matrix.
Motivated by GAE, lots of GNN-based graph representation learning methods have been proposed recently.
MGAE~\citep{wang2017mgae} presents a denoising marginalized autoencoder that reconstructs the node feature matrix $\mathbf{X}$.
ARGA~\citep{pan2018adversarially} adopts the adversarial learning strategy, and its generated node embeddings are forced to match a prior distribution.
DAEGC~\citep{wang2019attributed} exploits side information to generate node embeddings in a self-supervised way.
AGC~\citep{zhang2019attributed} proposes an improved filter matrix to better filter out the high-frequency noise. 
AGE~\citep{cui2020adaptive} further improves AGC by using the similarity of embedding rather than the adjacency matrix to consider original node feature information.
Compared with GNN-based graph representation learning methods that rely on the trainable parameters to learn node embeddings, our \sys is training-free and thus enjoys higher efficiency and scalability.

\section{Observation and Insight}
In this section, we make a quantitative analysis on the over-smoothing issue at the node level and then provide some insights when designing \sys on graphs.

\subsection{Feature Smoothing in Decoupled GNNs}
Recently, many works~\citep{wu2019simplifying, zhu2021simple,chen2020scalable, zhang2021graph} propose to decouple the feature smoothing and feature transformation in each GCN layer for scalable node classification.
Concretely, they execute the feature smoothing operation in advance, and the smoothed features are then fed into a simple MLP to generate the final predicted node labels.
The predictive node classification accuracy of these methods is comparable with or even higher than the one of coupled GNNs.
% and these works claim that the true power of GNNs lies in feature smoothing rather than feature transformation. 

These decoupled GNNs contain two parts: feature smoothing and MLP training. 
Feature smoothing aims to combine the graph structural information and node features into better features for the subsequent MLP;
while MLP training only takes in the smoothed feature and is specially trained for a given task.
As stated by previous decoupled GNNs~\citep{wu2019simplifying, zhu2021simple, zhang2021node}, the true success of GNNs lies in feature smoothing rather than feature transformation.
Correspondingly, we propose to remove feature transformation and preserve the key feature smoothing part alone for simple and scalable node representation.

There is another branch of GNNs that also decouple the feature smoothing and feature transformation.
The most representative method of this category is APPNP~\citep{klicpera2018predict}.
It first feeds the raw node features into an MLP to generate intermediate node embeddings. Then the personalized PageRank-based propagation operations are performed on the node embeddings to produce final prediction results.
However, compared with scalable decoupled GNNs mentioned in the previous paragraph, this branch of GNNs still have to recursively execute propagation operations in each training epoch, which makes it impossible to perform on large-scale graphs.
In the remaining part of this paper, the terminology ``decoupled GNNs'' refers particularly to the scalable decoupled GNNs mentioned in the previous two paragraphs.

\subsection{Measuring Smoothing Level}
To capture deep graph structural information, a straightforward way is to simply stack multiple GNN layers.
However, a large number of feature smoothing operations in a GNN model would lead to indistinguishable node embeddings, i.e., the over-smoothing issue~\citep{li2018deeper}.
Concretely, if we execute $\mathbf{\hat{A}}\mathbf{X}$ for infinite times, the node embeddings within the same connected component would reach a stationary state.
When adopting $\mathbf{\hat{A}}=\widetilde{\mathbf{D}}^{r-1}\tilde{\mathbf{A}}\widetilde{\mathbf{D}}^{-r}$, $\mathbf{\hat{A}}^{\infty}$ follows
 \begin{equation}
 \label{stationary}
 \small
 \hat{\mathbf{A}}^{\infty}_{i,j}  =  \frac{(d_i+1)^r(d_j+1)^{1-r}}{2m+n}.
 \end{equation}
which shows that the influence from node $v_{i}$ to $v_{j}$ is only determined by their degrees.
Under the extreme condition that $r=0$, all the nodes within one connected component have exactly the same representation, making it impossible to apply the node embeddings to subsequent tasks.

%\para{Over-smoothing Distance.} 
Here we introduce a new metric, ``Over-smoothing Distance'', to measure each node's smoothing level.
A smaller value indicates that the node is closer to the stationary state, i.e., closer to over-smoothing.
%In this paper, we introduce a new metric, ``Over-smoothing Distance'', to evaluate how close the aggregated state of each node is to its stationary state along the aggregation depth, and smaller distance indicates that the node state is closer to the stationary state.

%\sza{actually implement using $\mathbf{X}$ as reference, rather than $\hat{\mathbf{A}}^{\infty}\mathbf{X}$}
\begin{definition}[\textbf{Over-smoothing Distance}]
\label{df1}
The Over-smoothing Distance $D_{i}(k)$ parameterized by node $i$ and smoothing step $k$ is defined as
\begin{equation}
\small
\label{dik}
D_{i}(k) = Dis([\hat{\mathbf{A}}^{k}\mathbf{X}]_{i},[\hat{\mathbf{A}}^{\infty}\mathbf{X}]_{i}),
\end{equation}
where $[\hat{\mathbf{A}}^{k}\mathbf{X}]_{i}$ denotes the $i^{th}$ row of $\hat{\mathbf{A}}^{k}\mathbf{X}$, representing the representations of node $v_{i}$ after smoothing $k$ times; $[\hat{\mathbf{A}}^{\infty}\mathbf{X}]_{i}$ denotes the $i^{th}$ row of $\hat{\mathbf{A}}^{\infty}\mathbf{X}$, representing the stationary state of node $v_{i}$; $Dis(\cdot)$ is a distance function or a function positively relative with the difference, which can be implemented using Euclidean distance, the inverse of cosine similarity, etc.
\end{definition}

\subsection{Diverse Smoothing Speed across Nodes.}
\label{inconsistent}
\begin{figure}
% \vspace{-5mm}
	\centering
	\includegraphics[width=.7\linewidth]{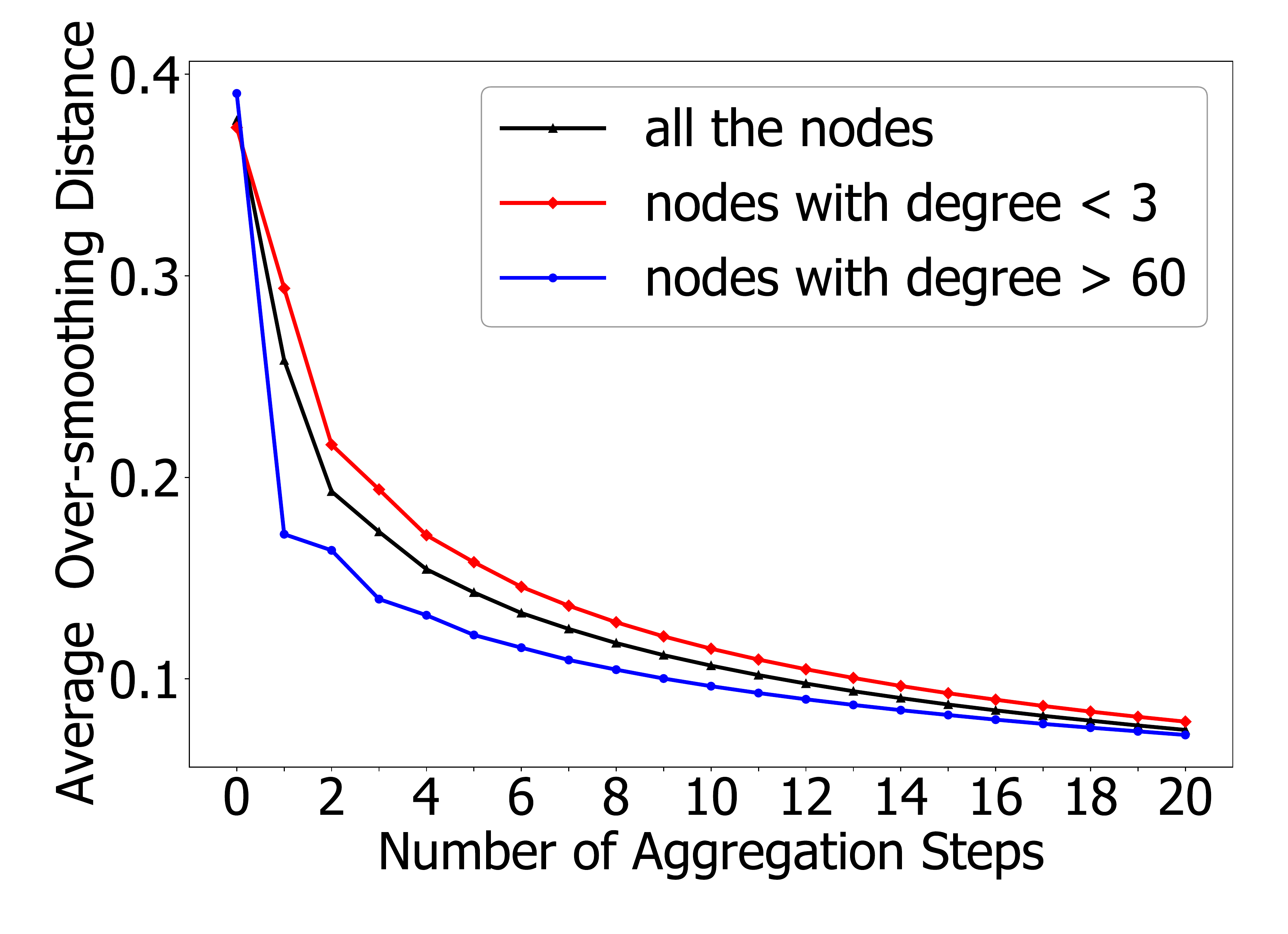}
%  	\vspace{-3mm}
	\caption{The diverse smoothing speed across nodes with different degrees. Nodes with larger degree have larger smoothing speed.}
	\label{smoothing_speed}
%  	\vspace{-2mm}
\end{figure}

To examine the factors that affect $D_i(k)$, we divide nodes in the PubMed dataset into three different groups according to their degrees.
In Figure~\ref{smoothing_speed}, we show the trends of nodes in the first group and the second group where the group-averaged $D_i(k)$ changes with the number of smoothing step increases.
The trend of $D_i(k)$ averaged over all the nodes is also provided, and the Euclidean distance is chosen as the distance function.
Figure~\ref{smoothing_speed} shows that the $D_i(k)$ of nodes with degrees larger than 60 drops more quickly than nodes with degrees smaller than 3, which implies that nodes with high degrees approach the stationary state more rapidly than the nodes with low degrees.

%The aggregation operation can be considered as a weighted average of the node embedding of all the nodes within its receptive field (i.e., the neighborhood can be reached).
%Fig.~\ref{RF} shows an example where two steps of feature aggregation have been carried out. 
%Obviously, the red node with large degree has much larger receptive field than the green node with small degree.
%Thus, nodes with large degrees should intuitively have smaller aggregation steps than nodes with small degrees to match their high receptive field expansion speed.

The quantitative analysis empirically illustrates that the degree of each node plays an essential role in one's optimal smoothing step.
Intuitively, nodes with high degrees should have relatively small smoothing steps than nodes with low degrees.
In addition, in Appendix~\ref{theory}, we have made a detailed theoretical analysis about the graph sparsity, another factor that influences the smoothing speed.

% \begin{figure}[tp!]
% \centering  
% \subfigure[diverse smoothing speed]{
% \label{smoothing_speed}
% \includegraphics[width=0.35\textwidth]{figure/smoothing_speed.pdf}}\hspace{5mm}
% \subfigure[an intuitive example]{
% \label{RF}
% \includegraphics[width=0.35\textwidth]{figure/RF_1.pdf}}
% \caption{Optimal aggregation steps analysis concerning the node degree. Receptive field is abbreviated as RF.}
% \label{fig.empirical}
% \end{figure}

\subsection{Design Insights of \sys}
Though using feature smoothing operations inside the previous decoupled GNNs is scalable for large graph representation learning, it will lead to sub-optimal node representation.
According to the observation in Sec.~\ref{inconsistent}, it is sub-optimal to execute feature smoothing for all the nodes indiscriminately as previous decoupled GNNs do since nodes with different structural properties have diverse smoothing speeds.
Therefore, node-adaptive feature smoothing (i.e., different nodes have different smoothing levels or mechanisms with equivalent effect) must be adopted to satisfy each node's diverse needs of smoothing level.
In our proposed method \sys, we utilize the metric $D_i(k)$ defined in Def.~\ref{df1} to help accomplish the aim of node-adaptive feature smoothing.

\section{Proposed method}
\label{sec4}
In this section, we present \sys, a training-free method for scalable graph representation learning.
We first compute the smoothed features with the feature smoothing operation. 
Then the feature ensemble operation is used to combine the smoothed features generated by different smoothing strategies. 
The pseudo code of \sys is provided in Appendix~\ref{details}. 
% In the following, we introduce our method in detail.

% \subsection{Two Feature Engineering Operations}
% Based on the above observation and insight, we propose a new graph representation learning framework \sys that consists of two novel feature engineering operations.

\subsection{Node-Adaptive Feature Smoothing}
\begin{figure}[tpb]
    \centering
    \includegraphics[width=0.95\linewidth]{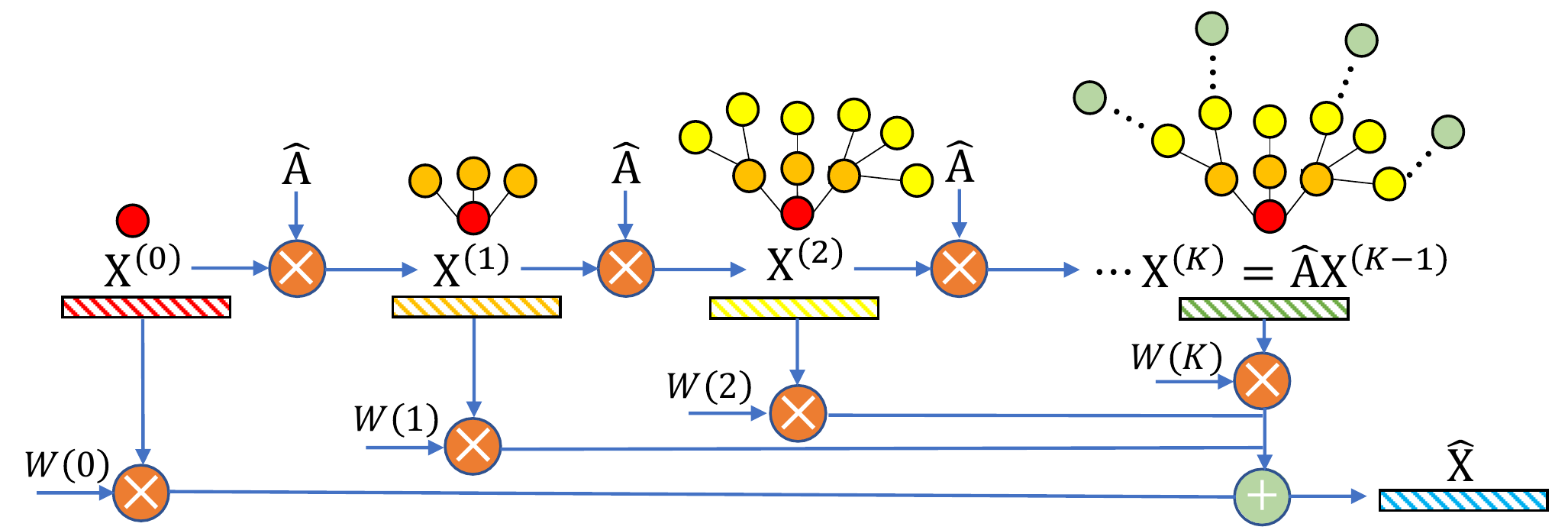}
    \caption{An overview of node-adaptive feature smoothing.}
    \label{FS}
    % \vspace{-4mm}
\end{figure}

Figure~\ref{FS} provides an overview of \sys. When repeatedly executing $\mathbf{X}^{(l)}= \hat{\mathbf{A}}\mathbf{X}^{(l-1)}$, the smoothed node embedding matrix $\mathbf{X}^{(l-1)}$ contains deeper graph structural information with $l$ increases.
The multi-scale node embedding matrices $\{\mathbf{X}^{(0)}, \mathbf{X}^{(1)}, ..., \mathbf{X}^{(K)}\}$ ($K$ is the maximal smoothing step) are then combined into a single matrix $\mathbf{\mathbf{\hat{X}}}$ such that both local and global neighborhood information are reserved.

% Previous works~\citep{zhu2021simple, chen2020scalable} also proposed similar approaches, but they regard all the nodes in the graph as a whole, neglecting the diversities across nodes.
The analysis in Sec.~\ref{inconsistent} illustrates that the speed each node achieves its stationary state is highly diverse, which suggests that nodes should be treated individually.
To this end, we define ``Smoothing Weight'' based on $D_i(k)$ introduced in Def.~\ref{df1} for each node so that the smoothing operation can be performed in a node-adaptive manner.

\begin{definition}[\textbf{Smoothing Weight}]
\label{df3}
The Smoothing Weight $w_{i}(k)$ parameterized by node $v_i$ and smoothing step $k$ is defined with the softmax value of 
\{$D_{i}(0)$, ${D_{i}(1)}$, $\cdots$, ${D_{i}(K)}$\}:
\begin{small}
\begin{equation}
\label{iw}
w_{i}(k) = e^{D_{i}(k)} / \sum \limits_{l=0}\limits^{K} e^{D_{i}(l)},
\end{equation}
\end{small}
where $K$ is the maximal smoothing step.
\end{definition}
% Larger $D_{i}(k)$ means that node $v_i$ is further from the stationary state, and $[\hat{\mathbf{A}}^{k}\mathbf{X}]_{i}$ empirically contains more relevant node information. 
% Therefore, for node $v_i$, aggregated feature with larger $D_{i}(k)$ (i.e., larger $w_i(k)$) should contribute more to the final node embedding.
To calculate $D_i(k)$ more efficiently, an alternative is to replace $[\hat{\mathbf{A}}^{\infty}\mathbf{X}]_{i}$ in Eq.~\ref{dik} with $\mathbf{X}_{i}$ and implement $Dis(\cdot)$ as the cosine similarity.
Larger $D_{i}(k)$ in this case means that node $v_i$ is farther from the stationary state and $[\hat{\mathbf{A}}^{k}\mathbf{X}]_{i}$ intuitively contains more relevant node information.
Therefore, for node $v_i$, the smoothed feature with larger $D_{i}(k)$ (i.e., larger $w_i(k)$) should contribute more to the final node embedding.
The smoothing weight can be formulated in the following matrix form:

\begin{definition}[\textbf{Smoothing Weight Matrix}]
The Smoothing Weight Matrix $\mathbf{W}(k)$ parameterized by smoothing step $k$ is defined as the diagonal matrix derived from $\eta(k) \in \mathbb{R}^n$:
\begin{small}
\begin{equation}
\label{w_k}
\mathbf{W}(k) = Diag(\eta(k)), \quad \eta(k)[i] = w_{i}(k), \quad \forall 1\le i\le n.
\end{equation}
\end{small}
% where $\eta_{k}$ is an $n$-dimension vector and 
% \begin{equation}
% \eta_{ki} = w_{i}(k),\quad\forall 1\le i\le n.
% \end{equation}
\end{definition}

Given the maximal smoothing step $K$, the multi-scale smoothed features $\{\mathbf{X}^{(0)}, \mathbf{X}^{(1)}, ..., \mathbf{X}^{(K)}\}$, and the corresponding Smoothing Weight Matrices $\{\mathbf{W}(0), \mathbf{W}(1), ..., \mathbf{W}(K)\}$ , the final smoothed feature $\mathbf{\hat{X}}$ can be represented as:
% \begin{small}
% \begin{equation}
% \label{smooth}
$\hat{\mathbf{X}} = \sum\limits_{k=0}\limits^{K}\mathbf{W}(k)\hat{\mathbf{A}}^{k}\mathbf{X}$.
% \end{equation}
% \end{small}
% where $\mathbf{X}$ is the initial input and $\mathbf{\hat{X}}$ is the output features after our graph feature engineering. 

%\sza{更多地从feature engineering的角度来讲，别从expressive power的角度讲}
\subsection{Feature Ensemble}
% \zwt{Mean,Max和Concat三种组合不同的r出来的smoothed feature的方法，参考graphsage}
%Previous work~\citep{hamilton2017inductive} has shown that different aggregators adopted in the feature aggregation operation lead to varied model capacity.
Different smoothing operators actually act as different knowledge extractors.
%, which capture information at different scales and dimensions.
For example, by setting $r = $ 0.5, 1 and 0, $\mathbf{\hat{A}}=\widetilde{\mathbf{D}}^{r-1}\widetilde{\mathbf{A}}\widetilde{\mathbf{D}}^{-r}$ represents the symmetric normalized adjacency matrix $\widetilde{\mathbf{D}}^{-1/2}\widetilde{\mathbf{A}}\widetilde{\mathbf{D}}^{-1/2}$~\citep{kipf2016semi}, the random walk transition probability matrix $\widetilde{\mathbf{D}}^{-1}\widetilde{\mathbf{A}}$~\citep{xu2018representation}, and the reverse random walk transition probability matrix $\widetilde{\mathbf{A}}\widetilde{\mathbf{D}}^{-1}$~\citep{zeng2019graphsaint}, respectively.
These variants of $\mathbf{\hat{A}}$ captures and reserves different scales and dimensions of knowledge from both graph structures and node features.

To achieve the same effect, the feature ensemble operation has multiple knowledge extractors: we vary the value of $r$ in the normalized adjacency matrix $\mathbf{\hat{A}}_r = \widetilde{\mathbf{D}}^{r-1}\widetilde{\mathbf{A}}\widetilde{\mathbf{D}}^{-r}$ to easily acquire different knowledge extractors.
These knowledge extractors are adopted inside the feature smoothing operation to generate different smoothed features.
Concretely, the value of $r$ controls the normalized weight of each edge. 
So different values of $r$ generate different weight values for all the edges in the graph, which would increase the diversity of our smoothed features evidently.
The ablation study in Sec.~\ref{ablation} shows that the increased diversity contributes a lot to the high performance of our generated node embeddings when applied to downstream tasks.
%With the smoothed features under different $r$, we propose to combine them because a single feature aggregation strategy with a fixed $r$ might be sub-optimal in extracting the knowledge of graph structure and node features.

The detailed procedure of feature ensemble operation is as follows:
Given $\{r_1, r_2, ..., r_T\}$, we firstly perform the feature smoothing operation to generate corresponding smoothed features $\{\hat{\mathbf{X}}^{(1)}, \hat{\mathbf{X}}^{(2)}, ..., \hat{\mathbf{X}}^{(T)}\}$. 
Then, we combine them as
% \begin{small}
% \begin{equation}
% \label{ens}
    $\mathbf{Z}  \gets \oplus_{i\in \{1, 2, ..., T\} }\hat{\mathbf{X}}^{(i)}$,
% \end{equation}
% \end{small}
where $\oplus$ is the ensemble strategy, which can be implemented as concatenating, mean pooling, max pooling, etc.

\subsection{Theoretical Analysis}
The essential kernel of \sys is the Smoothing Weight, which determines the output results. We now analyze the factors affecting the value of Smoothing Weight. To simplify our analysis, we suppose $r=0$ in the normalized adjacency matrix and apply Euclidean distance as the distance function in Definition \ref{df1}. 
Thus we have
\begin{equation}
\small
\label{eq17}
D_{i}(k) = ||[\hat{\mathbf{A}}^{k}\mathbf{X}]_{i}-[\hat{\mathbf{A}}^{\infty}\mathbf{X}]_{i}||_{2},
\end{equation}
where $||\cdot||_{2}$ symbols two-norm.
\begin{theorem}
For any node $i$ in graph $\mathcal{G}$, there always exists
\begin{equation}
\small
D_{i}(k) \le \lambda_{2}^{k}\sqrt{\frac{\sum\limits_{j=1}\limits^{n}(\tilde{d_{j}}||\mathbf{X}_{j}||_{2}^{2})}{\tilde{d_{i}}}},
\label{thm}
\end{equation}
where $\tilde{d_{i}}$ = $d_{i}+1$, $||\mathbf{X}_{j}||_{2}$ denotes the two-norm of the initial feature of node $j$, and $0<\lambda_{2}<1$ denotes the second largest eigenvalue of the normalized adjacency matrix $\hat{\mathbf{A}}$. 
\end{theorem}

Eq. \ref{thm} shows that the nodes with smaller degrees may have larger $D_{i}(k)$. Combined with definition \ref{df3}, we infer that larger $D_{i}(k)$ makes the $\max_{k}D_{i}(k)$ more dominant after the softmax operation, causing that weighted average results depend more on itself and its near neighbors. 
Inversely, for the nodes with smaller degrees, its result of weighted average depends more equally on all itself, its near neighbors, and its distant neighbors.

Besides, $D_{i}(k)$ in a sparser graph ($\lambda_{2}$ is positively relative with the sparsity of a graph) decays slower as $k$ increases. Thus, the weighted average result depends more equally on itself, its near neighbors and its distant neighbors. While for the nodes in a denser graph, the weighted average result depends more on its near neighbors and itself.

The detailed proof of Eq.~\ref{thm} is in Appendix~\ref{t1}. Besides, we theoretically analyze how \sys prevent over-smoothing in Appendix~\ref{t2}. Lastly, we also theoretically show how \sys leverages the multiple features over different smooth steps in a node-adaptive manner in Appendix~\ref{t3}.

% \subsection{\sys Pipeline}
% Alg.~\ref{alg:graphfe} shows the whole pipeline of our proposed \sys. 
% We first initialize $\mathbf{X}^{(0)}$ as the original feature matrix $\mathbf{X}$. 
% Given the normalization parameter $r_t$, we obtained the corresponding normalized adjacency matrix $\mathbf{\hat{A}}_{r_t} = \widetilde{\mathbf{D}}^{r_t-1}\widetilde{\mathbf{A}}\widetilde{\mathbf{D}}^{-r_t}$, which acts as knowledge extractors (line 4). 
% After that, for each node, we use E.q.~\ref{dik} to calculate the Over-smoothing Distance with all the $k$ ranging from $0$ to $K$ (line 6, 7). 
% Then we calculate its Aggregation Weights through E.q.~\ref{iw} (line 8, 9). 
% After obtaining Aggregation Weights with all $k$ and $i$, we construct the Aggregation Weight matrix for each $k$ with E.q.~\ref{w_k} (line 10, 11). 
% Next, we compute the NAFS output $\hat{\mathbf{X}}^{(t)}$ with $\hat{\mathbf{X}}^{(t)} = \sum\limits_{k=0}\limits^{K}\mathbf{W}(k)\hat{\mathbf{A}}_{r_t}^{k}\mathbf{X}$ (line 12). 
% Finally, we compute the final embedding result of all $t$ through $\mathbf{Z}  \gets \oplus_{i\in \{1, 2, ..., T\} }\hat{\mathbf{X}}^{(i)}$ (line 14).

\section{Advantages over Traditional Approaches}

% \textbf{Efficiency Optimization}
% \zwt{不用一个section，说一下当feature维度很大的时候，可以先用AutoEncoder或者LDA降维}
\sys generates node embeddings in a training-free manner, making it highly efficient and scalable.
Moreover, the node-adaptive smoothing strategy enables it to capture deep structural information.
%As mentioned in Sec.~\ref{GAE_related}, many GNN-based graph representation learning methods have been proposed after GAE.
%Although they more or less have alleviated some flaws of GAE, most of them still need to reconstruct the adjacency matrix to compute the training loss, making them as slow and unscalable as GAE.
In this subsection, we analyze the advantages of our \sys over GAE and its variants.

\para{Deep Structural Information.}
%\zwt{能很好利用local和global的图结构信息，即使做到10000层效果依旧强劲.1.性能远远：避免oversmooth 2.训练的原因：我们的时间和空间复杂度低：单机，对分布式支持很好}
%Apart from the training cost, our proposed NAFS can effectively take advantage of deep structural information.
By assigning each node with personalized smoothing weights, \sys can gather deep structural information without encountering the over-smoothing issue and keep the time and memory cost low.
While for GAE and its variants, they either 1) have a coupled structure that cannot go deep due to low efficiency and high memory cost (e.g., GAE) or 2) are unable to adaptively capture structural information and encounter the over-smoothing issue when going deeper (e.g., AGE).

% \zwt{给个table，show一下memory和complexity等}\\
\para{Efficiency.}
%\zwt{速度超快，几个数量级的快}
Compared with GAE and its variants, our proposed \sys does not have any trainable parameters, giving it a significant advantage in efficiency.
%Denote by $n$ the number of nodes, $f$ the number of attributes, $k$ the aggregation steps, and m the number of nonzero entries of the sparse adjacency matrix $A$.
When generating node embeddings, \sys only needs to execute the feature smoothing and feature ensemble operations, which has a time complexity of $\mathcal{O}(Kmf+Kn)$, where $K$ denotes the maximal smoothing step.
On the other hand, the asymptotic training time complexity of GAE and its variants is $\mathcal{O}(nf^2+n^2f)$, which is one magnitude larger than \sys.
In Sec.~\ref{effi_compare}, we further present efficiency comparison in detail on real-world datasets between \sys and GAE.

\para{Memory Cost.}
% \zwt{分布式}
% \noindent\textbf{Low Memory Usage}
%\zwt{1.单机内存占用小，单机可以支持千亿节点的大图训，空间复杂度低}
%\zwt{2.分布式开销小，不需要重复拉feature}
Our proposed \sys enjoys not only high efficiency but also low memory cost.
\sys only requires to store the sparse adjacency matrix and the smoothed features, and thus the memory cost is $\mathcal{O}(m+nf)$, which grows linearly with graph size $n$ in typical real-world graphs.
For GAE and its variants, they need additional space to store the training parameters and especially the reconstructed dense adjacency matrix, making their memory cost $\mathcal{O}(n^2+m+nf)$, which is also one magnitude larger than \sys.
Scalability comparison conducted on synthetic datasets between \sys and GAE can be found in Appendix~\ref{scalability_comp}.

\section{Experimental Results}
\label{exp_sec}
In this section, we conduct extensive experiments to evaluate the proposed \sys.
We first introduce the considered baseline methods, used datasets, and experimental settings.
Then, we demonstrate the advantages of \sys from the following three perspectives: (1) end-to-end comparison with the state-of-the-art methods, (2) scalability and efficiency, and (3) effectiveness along with ablation studies.

%\subsection{Downstream Tasks}
% \zwt{在三个task上怎么搞}
% \noindent\textbf{Node Clustering.}
% For the node clustering task, we apply K-Means~\citep{hartigan1979algorithm} to the node embeddings to get the clustering results.
% Three widely-used metrics are used for evaluation: Accuracy (ACC), Normalized Mutual Information (NMI), and Adjusted Rand Index (ARI).

% \noindent\textbf{Link Prediction.}
% For the link prediction task, 5\% and 10\% edges are randomly reserved for the validation set and the test set.
% During the aggregation, the normalized adjacency matrix $\hat{A}$ containing only 85\% edges is used.
% Once the graph embedding has been generated, we follow the same procedure as GAE to reconstruct the adjacency matrix.
% Two metrics: Area Under Curve (AUC) and Average Precision (AP) are used to evaluate the performance of the link prediction task.

\subsection{Datasets and Baseline Methods}
%\para{Datasets.}
\label{sec:datasets}
Several widely-used network datasets (i.e., Cora, Citeseer, PubMed, Wiki and ogbn-products) are used in our experiments. 
% In Cora and Citeseer, the node features are binary word vectors; while in PubMed and Wiki, each node has a TF-IDF weighted word vector.
We include the properties of these datasets in Appendix~\ref{data}.
%Besides, considering the prevalence of Open Graph Benchmark (OGB)~\citep{hu2020ogb} dataset, we also test node classifcation task on three large ogbn datasets: ogbn-arxiv, ogbn-products and ogbn-papers100M.
%It is worth noting that the ogbn-papers100M dataset contains more than 100 million nodes and 1 billion edges, which can be used as a criterion for whether a method is scalable.

%\para{Baselines.} 
For different tasks, the compared baselines are as follows:
%To evaluate the performance of NAFS, we compare our method with the following representative methods:
\begin{itemize}
    \item \textbf{ Link prediction:} 
    Spectral Clustering (SC)~\citep{ng2002spectral}, DeepWalk~\citep{perozzi2014deepwalk}, GAE and VGAE~\citep{kipf2016variational}, ARGA and ARVGA~\citep{pan2018adversarially}, GALA~\citep{park2019symmetric}, and AGE~\citep{cui2020adaptive}.
    
    \item \textbf{ Node clustering:} GAE and VGAE~\citep{kipf2016variational}, MGAE~\citep{wang2017mgae}, ARGA and ARVGA~\citep{pan2018adversarially},  AGC~\citep{zhang2019attributed}, DAEGC~\citep{wang2019attributed}, and AGE~\citep{cui2020adaptive}.
\end{itemize}
For \sys, we investigate the following three variants: NAFS-mean, NAFS-max, and NAFS-concat.
They all have the complete NAFS framework and only differ in the ensemble strategy adopted in feature ensemble operation.

\begin{table}[t]
\centering
    \captionof{table}{Link prediction performance comparison.} 
    % \vspace{-2mm}
    \resizebox{0.95\linewidth}{!}{
    \begin{tabular}{c|cc|cc|cc}
    \toprule
    \multirow{2}{*}{\textbf{Methods}} & \multicolumn{2}{c|}{\textbf{Cora}} & \multicolumn{2}{c|}{\textbf{Citeseer}} & \multicolumn{2}{c}{\textbf{PubMed}} \\ \cline{2-7} 
                             & \textbf{AUC}    & \textbf{AP}    & \textbf{AUC}    & \textbf{AP}      & \textbf{AUC}     & \textbf{AP}\\ 
    \midrule
    SC& 84.6$\pm$0.0 & 88.5$\pm$0.0 & 80.5$\pm$0.0  & 85.0$\pm$0.0  & 84.2$\pm$0.0  & 87.8$\pm$0.0  \\
    DeepWalk& 83.1$\pm$0.3 & 85.0$\pm$0.4  & 80.5$\pm$0.5 & 83.6$\pm$0.4 & 84.4$\pm$0.4 & 84.1$\pm$0.5  \\
    GAE & 91.0$\pm$0.5 & 92.0$\pm$0.4 & 89.5$\pm$0.3 & 89.9$\pm$0.4 & 96.4$\pm$0.4 & 96.5$\pm$0.5 \\
    VGAE & 91.4$\pm$0.5 & 92.6$\pm$0.4 & 90.8$\pm$0.4  & 92.0$\pm$0.3  & 94.4$\pm$0.5 & 94.7$\pm$0.4  \\
    ARGA & 92.4$\pm$0.4 & 93.2$\pm$0.3 & 91.9$\pm$0.5 & 93.0$\pm$0.4 & 96.8$\pm$0.3 & 97.1$\pm$0.5 \\
    ARVGA &92.4$\pm$0.4 &92.6$\pm$0.4 & 92.4$\pm$0.5 & 93.0$\pm$0.3 & 96.5$\pm$0.5 & 96.8$\pm$0.4\\
    %SIGN &88.6 &90.1 &86.3&87.7 &93.5 &94.5 \\
    GALA &92.1$\pm$0.3 &92.2$\pm$0.4 & 94.4$\pm$0.5 & 94.8$\pm$0.5 & 93.5$\pm$0.4 & 94.5$\pm$0.4 \\
    AGE &\textbf{95.1$\pm$0.5} & \textbf{94.6$\pm$0.5} & \textbf{96.3$\pm$0.4} & \textbf{96.6$\pm$0.4} & 94.3$\pm$0.3 & 93.5$\pm$0.5 \\
    \midrule
    % NAFS-simple & 91.9$\pm$0.0 & 93.1$\pm$0.0 & 94.1$\pm$0.0 & 95.2$\pm$0.0 & 97.4$\pm$0.0 & \textbf{97.2$\pm$0.0} \\
    NAFS-mean & 92.6$\pm$0.0 & 93.9$\pm$0.0 & \underline{94.9$\pm$0.0} & 95.9$\pm$0.0 & 97.4$\pm$0.0 & \textbf{97.2$\pm$0.0} \\
    NAFS-max &\underline{93.0$\pm$0.0} &\underline{94.2$\pm$0.0} & 94.8$\pm$0.0 &\underline{96.0$\pm$0.0} &\underline{97.5$\pm$0.0} &\underline{97.1$\pm$0.0} \\
    NAFS-concat &92.6$\pm$0.0 &93.8$\pm$0.0 &93.7$\pm$0.0 &93.1$\pm$0.0 &\textbf{97.6$\pm$0.0} &\textbf{97.2$\pm$0.0} \\
    \bottomrule
    \end{tabular}
    }
    % \vspace{-4mm}
    \label{link_pre_performance}
\end{table}

\subsection{Experimental Settings}
\label{sec:setups}
\para{Node Clustering.}
For the node clustering task, we apply K-Means~\citep{hartigan1979algorithm} to node embeddings to get the clustering results.
Three widely-used metrics are used for evaluation: Accuracy (ACC), Normalized Mutual Information (NMI), and Adjusted Rand Index (ARI).

\para{Link Prediction.}
For the link prediction task, 5\% and 10\% edges are randomly reserved for the validation set and the test set.
%During the aggregation, the normalized adjacency matrix $\mathbf{\hat{A}}$ containing only 85\% edges is used.
Once the node embeddings have been generated, we follow the same procedure as GAE to reconstruct the adjacency matrix.
Two metrics - Area Under Curve (AUC) and Average Precision (AP) are used in the evaluation of the link prediction task.

We run the compared baselines with 200 epochs and repeat the experiment 10 times on all the datasets, and report the mean value of each evaluation metric. 
The detailed setting of the hyperparameters and experimental environment are introduced in Appendix~\ref{HPO} and~\ref{envi}.

\begin{table*}[tbp]
\centering
{
\noindent
\caption{Node clustering performance comparison.}
\label{cluster_performance}
\renewcommand{\multirowsetup}{\centering}
\resizebox{0.95\linewidth}{!}{
\begin{tabular}{c|ccc|ccc|ccc|ccc}
\toprule
\multirow{2}{*}{\textbf{Methods}} & \multicolumn{3}{c|}{\textbf{Cora}} & \multicolumn{3}{c|}{\textbf{Citeseer}} & \multicolumn{3}{c|}{\textbf{PubMed}} & \multicolumn{3}{c}{\textbf{Wiki}}\\ \cline{2-13} 
& \textbf{ACC}    & \textbf{NMI}    & \textbf{ARI}    & \textbf{ACC}      & \textbf{NMI}     & \textbf{ARI}     & \textbf{ACC}     & \textbf{NMI}     & \textbf{ARI} & \textbf{ACC}     & \textbf{NMI}     & \textbf{ARI} \\ 
\midrule
GAE & 53.3$\pm$0.2  & 40.7$\pm$0.3  & 30.5$\pm$0.2  & 41.3$\pm$0.4    & 18.3$\pm$0.3   & 19.1$\pm$0.3   & 63.1$\pm$0.4   & 24.9$\pm$0.3   & 21.7$\pm$0.2 & 37.9$\pm$0.2 & 34.5$\pm$0.3 & 18.9$\pm$0.2\\
VGAE & 56.0$\pm$0.3  & 38.5$\pm$0.4  & 34.7$\pm$0.3  & 44.4$\pm$0.2    & 22.7$\pm$0.3   & 20.6$\pm$0.3   & 65.5$\pm$0.2   & 25.0$\pm$0.4   & 20.3$\pm$0.2  & 45.1$\pm$0.4 & 46.8$\pm$0.3 & 26.3$\pm$0.4\\
MGAE & 63.4$\pm$0.5  & 45.6$\pm$0.3  & 43.6$\pm$0.4  & 63.5$\pm$0.4    & 39.7$\pm$0.4   & 42.5$\pm$0.5   & 59.3$\pm$0.5   & 28.2$\pm$0.2   & 24.8$\pm$0.4  & 52.9$\pm$0.3 & \underline{51.0$\pm$0.4} & \textbf{37.9$\pm$0.5}\\
ARGA & 63.9$\pm$0.4  & 45.1$\pm$0.3  & 35.1$\pm$0.5  & 57.3$\pm$0.5    & 35.2$\pm$0.3   & 34.0$\pm$0.4   & 68.0$\pm$0.5   & 27.6$\pm$0.4   & 29.0$\pm$0.4  & 38.1$\pm$0.5 & 34.5$\pm$0.3 & 11.2$\pm$0.4\\
ARVGA & 64.0$\pm$0.5 & 44.9$\pm$0.4  & 37.4$\pm$0.5  & 54.4$\pm$0.5    & 25.9$\pm$0.5   & 24.5$\pm$0.3   & 51.3$\pm$0.4   & 11.7$\pm$0.3   & 7.8$\pm$0.2  & 38.7$\pm$0.4 & 33.9$\pm$0.4 & 10.7$\pm$0.2\\
AGC & 68.9$\pm$0.5  & 53.7$\pm$0.3  & 48.6$\pm$0.3  & 66.9$\pm$0.5    & 41.1$\pm$0.4   & 41.9$\pm$0.5   & 69.8$\pm$0.4   & 31.6$\pm$0.3   & 31.8$\pm$0.4  & 47.7$\pm$0.3 & 45.3$\pm$0.5 & 34.3$\pm$0.4\\
DAEGC & 70.2$\pm$0.4  & 52.6$\pm$0.3  & 49.7$\pm$0.4  & 67.2$\pm$0.5    & 39.7$\pm$0.5   & 41.1$\pm$0.4   & 66.8$\pm$0.5   & 26.6$\pm$0.2   & 27.7$\pm$0.3  & 48.2$\pm$0.4 & 44.8$\pm$0.4 & 33.1$\pm$0.3\\
AGE & \underline{72.8$\pm$0.5}  & \underline{58.1$\pm$0.6}  & \textbf{56.3$\pm$0.4}  & 70.0$\pm$0.3    & 44.6$\pm$0.4   & 45.4$\pm$0.5   & 69.9$\pm$0.5   & 30.1$\pm$0.4   & 31.4$\pm$0.6  & 51.1$\pm$0.6 & \textbf{53.9$\pm$0.4} & \underline{36.4$\pm$0.5}\\
\midrule
% NAFS-simple & 66.4$\pm$0.0 & 50.8$\pm$0.0 & 41.8$\pm$0.0 & \underline{71.5$\pm$0.0} & 44.4$\pm$0.0 & \underline{46.8$\pm$0.0} & 63.0$\pm$0.0 & 27.9$\pm$0.0 & 25.1$\pm$0.0 & 44.8$\pm$0.0 & 43.1$\pm$0.0 & 20.0$\pm$0.0\\
NAFS-mean & 70.4$\pm$0.0 & 56.6$\pm$0.0 & 48.0$\pm$0.0 & \textbf{71.8$\pm$0.0} & \underline{45.1$\pm$0.0} & \textbf{47.6$\pm$0.0} & \underline{70.5$\pm$0.0} & \textbf{33.9$\pm$0.0}  & \textbf{33.2$\pm$0.0} & \textbf{54.6$\pm$0.0} & 49.4$\pm$0.0 & 27.3$\pm$0.0\\
NAFS-max & 70.8$\pm$0.0  & 56.6$\pm$0.0  & 49.0$\pm$0.0  & 70.1$\pm$0.0 & \underline{45.1$\pm$0.0}   & 44.7$\pm$0.0 & \textbf{70.6$\pm$0.0} & \underline{33.4$\pm$0.0} & \underline{33.1$\pm$0.0} & 51.4$\pm$0.0 & 45.8$\pm$0.0 & 25.5$\pm$0.0\\
NAFS-concat & \textbf{75.4$\pm$0.0} & \textbf{58.6$\pm$0.0} & \underline{53.8$\pm$0.0}  & \underline{71.1$\pm$0.0} & \textbf{45.8$\pm$0.0} &   \underline{46.1$\pm$0.0} & \underline{70.5$\pm$0.0} & \textbf{33.9$\pm$0.0} & \textbf{33.2$\pm$0.0} & \underline{53.6$\pm$0.0} & 50.5$\pm$0.0 & 26.3$\pm$0.0 \\
\bottomrule
\end{tabular}}}
% \vspace{-1mm}
\end{table*}

\begin{figure*}[t]
  \begin{minipage}[b]{0.6\textwidth}
  \centering  
\subfigure[Weighting strategies]{
\label{ablation_weight}
\includegraphics[width=0.47\textwidth]{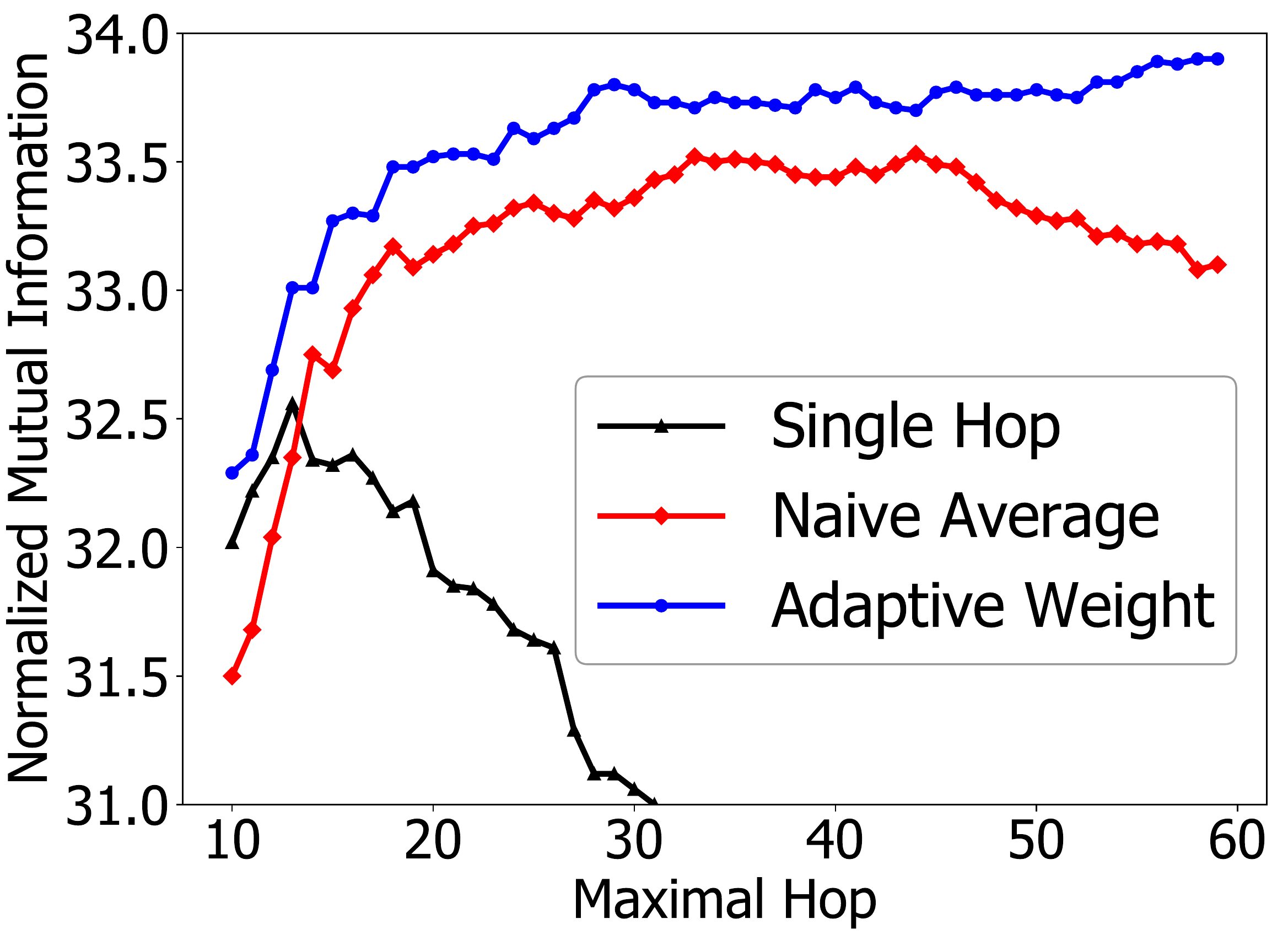}}\hspace{0.5mm}
\subfigure[Ensemble strategies]{
\label{ablation_ensemble}
\includegraphics[width=0.49\textwidth]{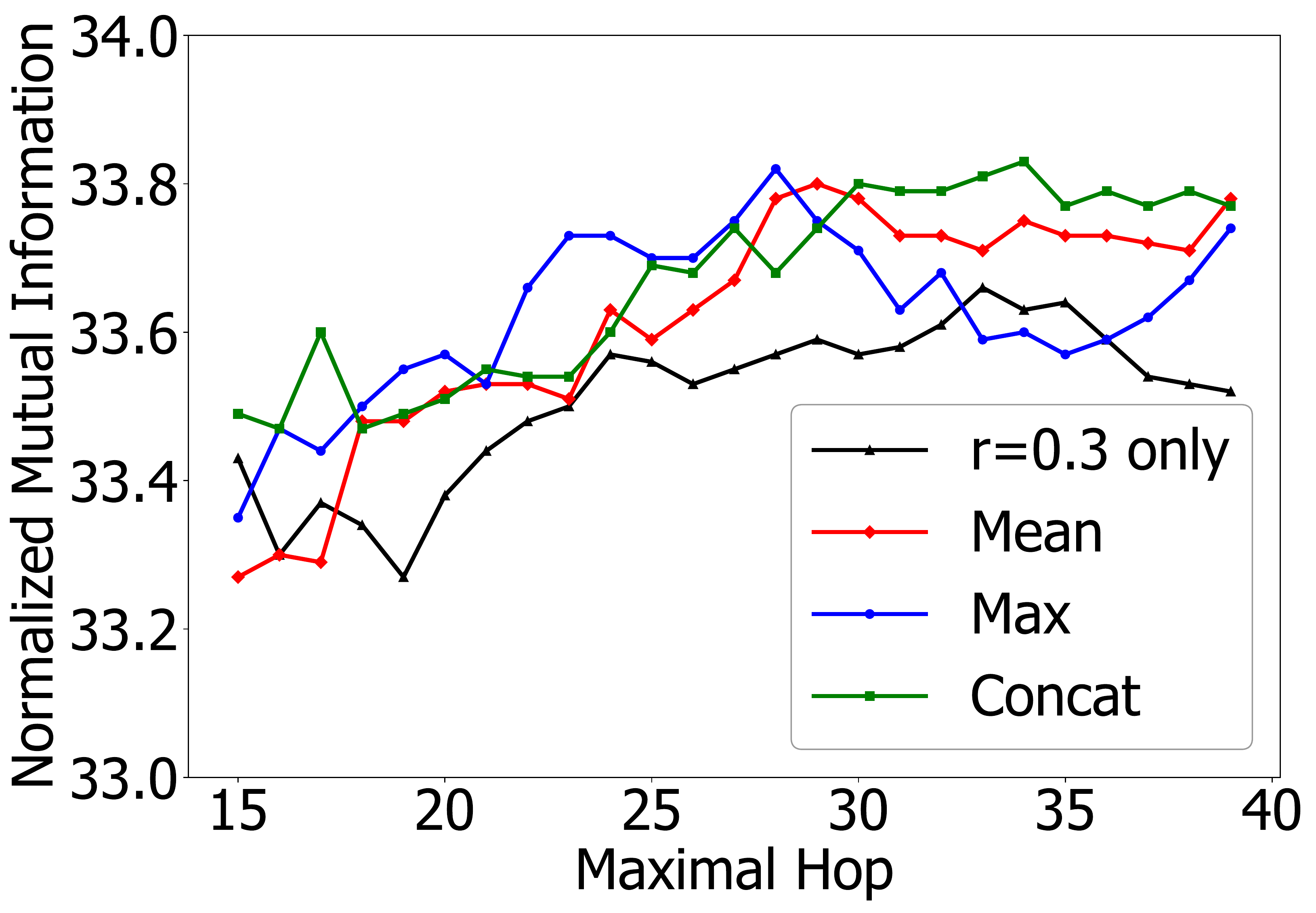}}
\caption{Ablation study to verify the effectiveness of each strategy.}
\label{fig.ablation}
    \end{minipage}
    \hspace{0.5mm}
  \begin{minipage}[b]{0.35\textwidth}
    \centering
    \scalebox{0.15}{
     \includegraphics{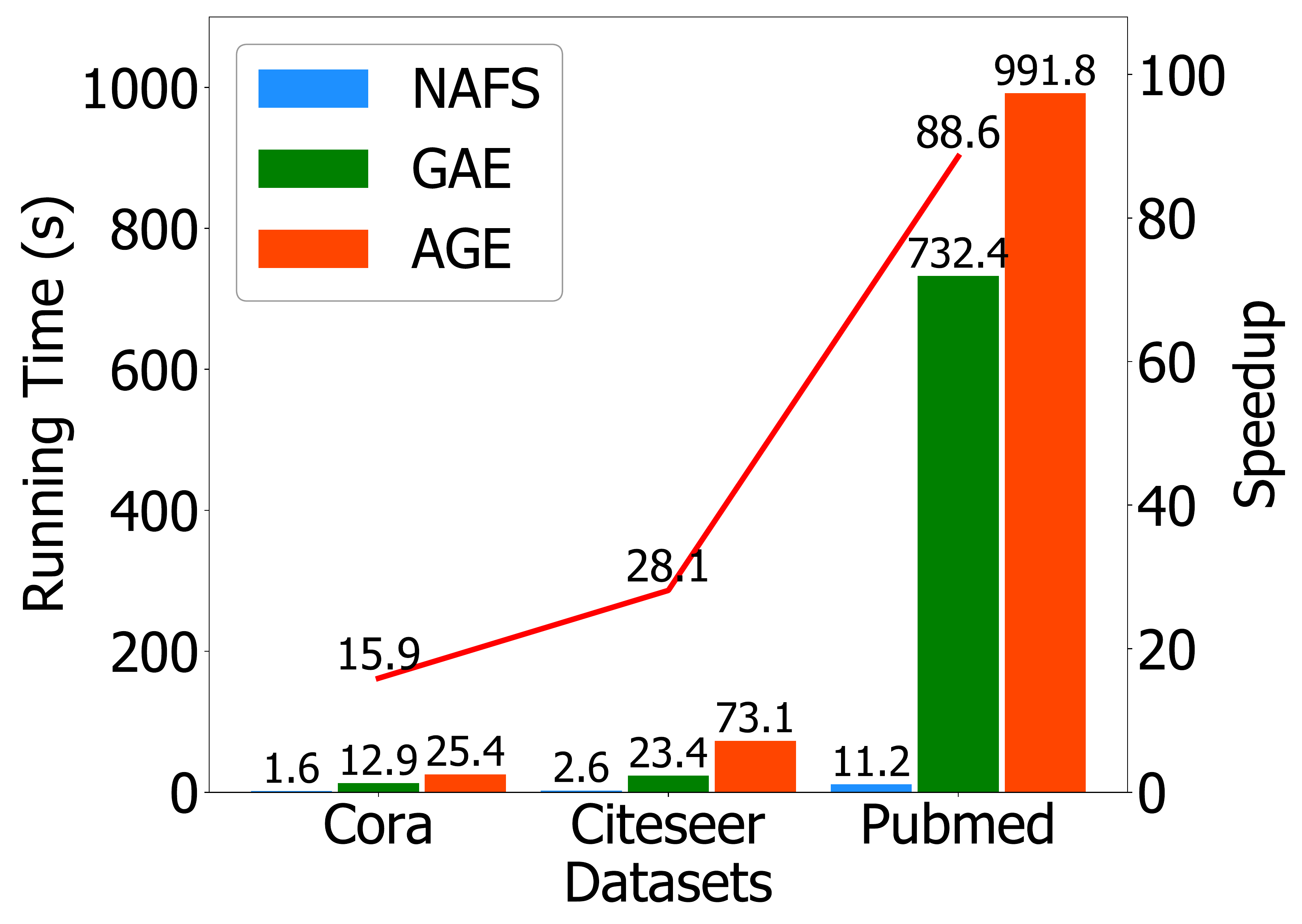}}
    \captionof{figure}{Efficiency comparison on the three citation networks.}
    \label{efficiency}
  \end{minipage}
%   \hfill
  \end{figure*}
  
\subsection{End-to-end Comparison} 
\para{Link Prediction.}
Table~\ref{link_pre_performance} shows the performance of different methods on the link prediction task.
On the three datasets, the proposed \sys consistently achieves the best or the second-best performance compared with all the baseline methods.
Remarkably, NAFS-concat achieves state-of-the-art performance on PubMed, and outperforms the current state-of-the-art method - ARGA by 0.8\% and 0.1\% on AUC and AP, respectively.

\para{Node Clustering.}
The node clustering results of each method are shown in Table~\ref{cluster_performance}.
Among three ensemble strategies, NAFS-concat has the overall best performance across the three datasets, which also consistently outperforms the strongest baseline - AGE.
For example, the best of NAFS variants exceeds AGE by 2.6\%, 1.8\%, 0.7\%, and 3.5\% on Cora, Citeseer, PubMed, and Wiki, respectively.
Although NAFS-concat has the overall best performance, it falls behind NAFS-mean or NAFS-max on some metrics of some datasets; and it needs more CPU memory to store the node embeddings and longer time for inference.

It is quite interesting to find that the three NAFS variants outperform training-based GAE on all the four datasets, and they outperforms the current SOTA method, AGE, in most cases.
The competitive performance of NAFS further illustrates that it is possible to achieve decent performance on graphs with only feature smoothing without any training parameters.

\subsection{Efficiency Analysis}
\label{effi_compare}
To validate the efficiency of \sys, we compare it with GAE and AGE, measuring their overall running time for generating node embeddings.
The comparison is conducted on the three citation networks.
For fairness, all methods only use CPUs for computation.
Figure~\ref{efficiency} showcases the results along with the speedup ratio of NAFS against AGE.

Figure~\ref{efficiency} shows that \sys is significantly faster than the considered methods on the three datasets. 
For example, NAFS is almost two magnitudes faster than AGE on the relatively large dataset - PubMed.
The high efficiency of NAFS is attributed to no trainable parameters, while GAE and AGE are both training-based methods.
Although AGE executes the feature smoothing step in advance, it adopts a time-consuming ranking policy to select positive and negative examples.
This alteration in AGE brings both superior performance and longer running time compared with GAE.

\begin{figure*}[htbp]
\centering  
\subfigure[original]{
\includegraphics[width=0.12\textwidth]{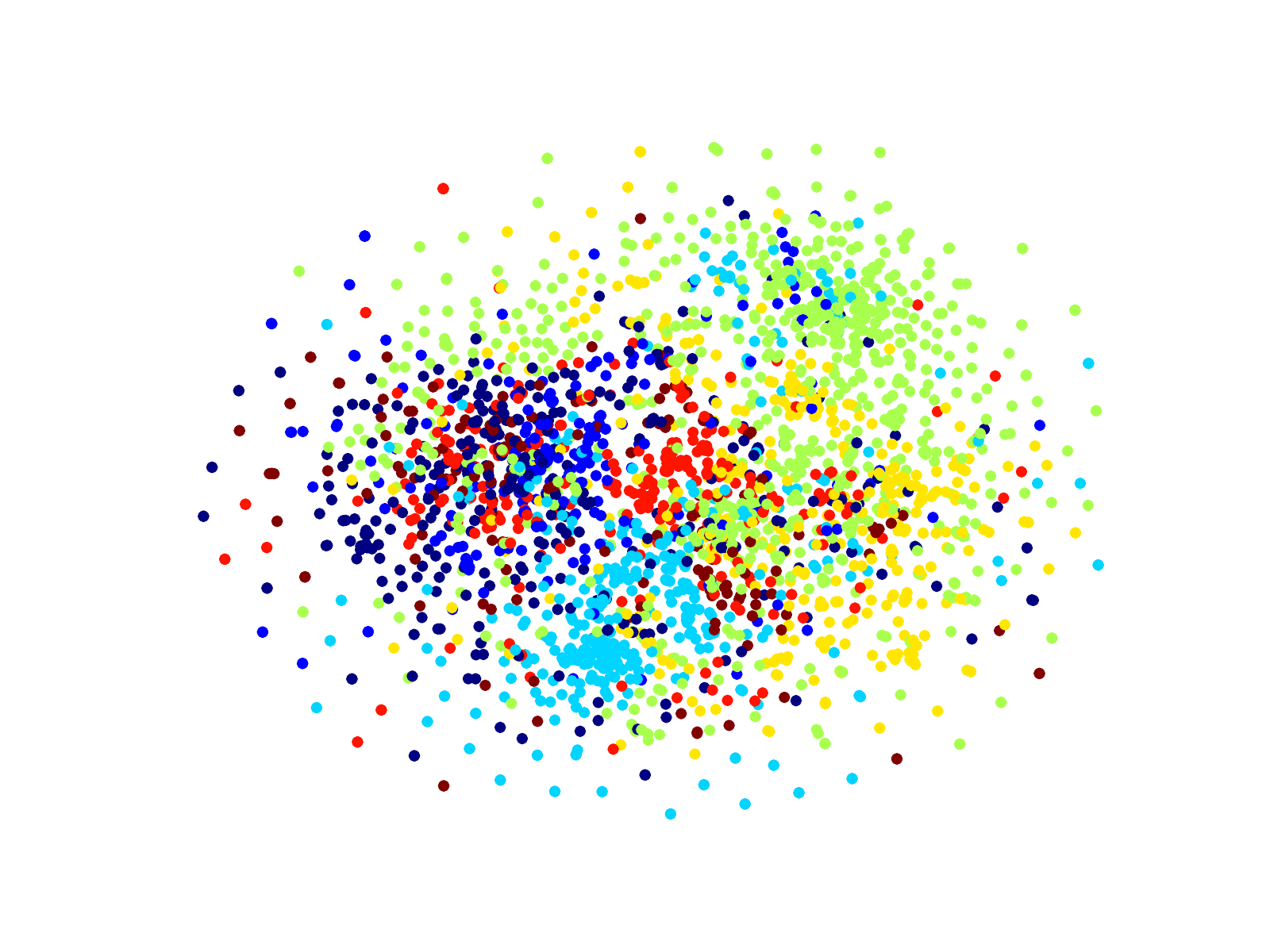}}\hspace{-1mm}
\subfigure[our 5]{
\includegraphics[width=0.12\textwidth]{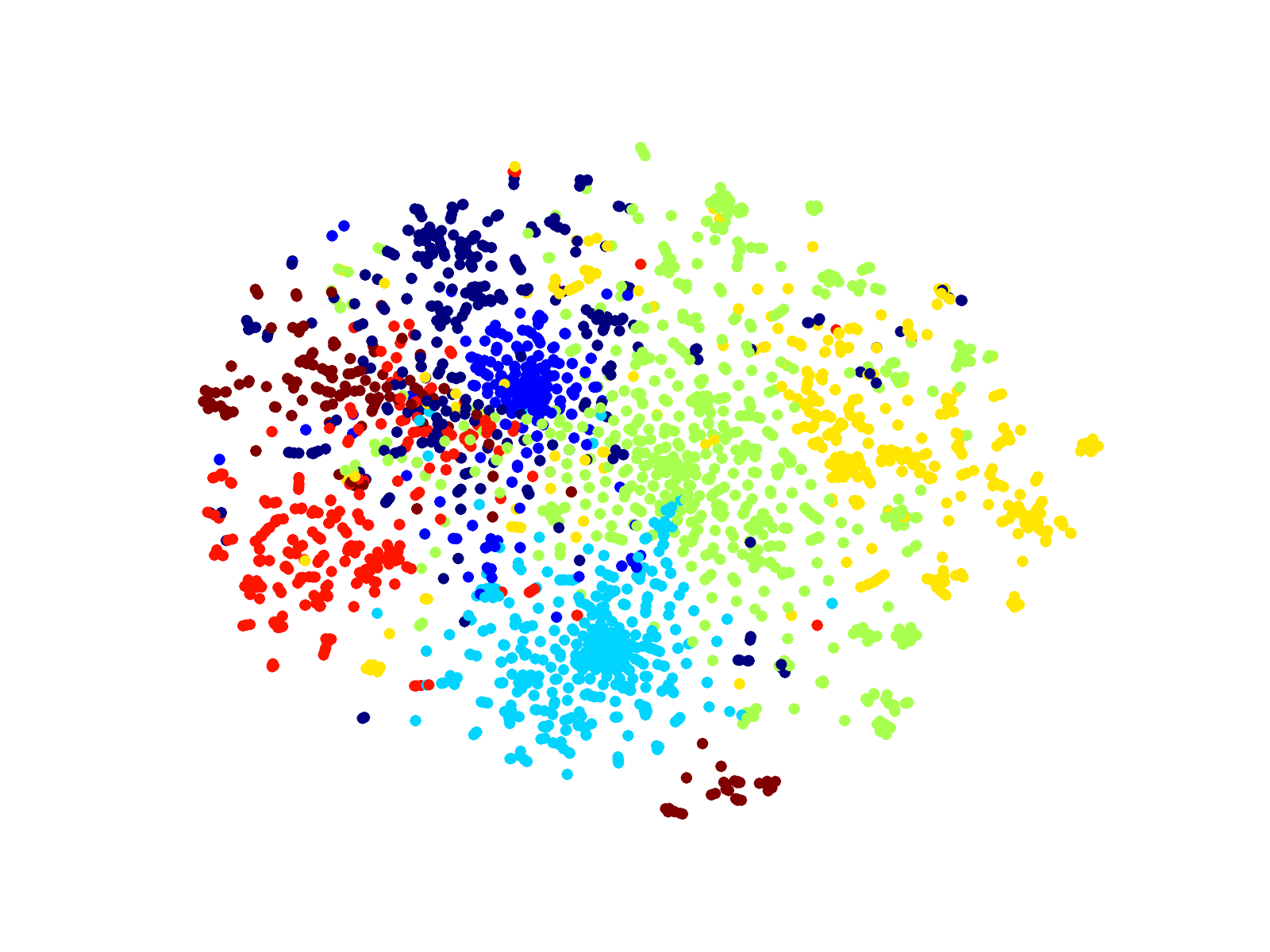}}\hspace{-1mm}
\subfigure[our 10]{
\label{our_hop_10}
\includegraphics[width=0.12\textwidth]{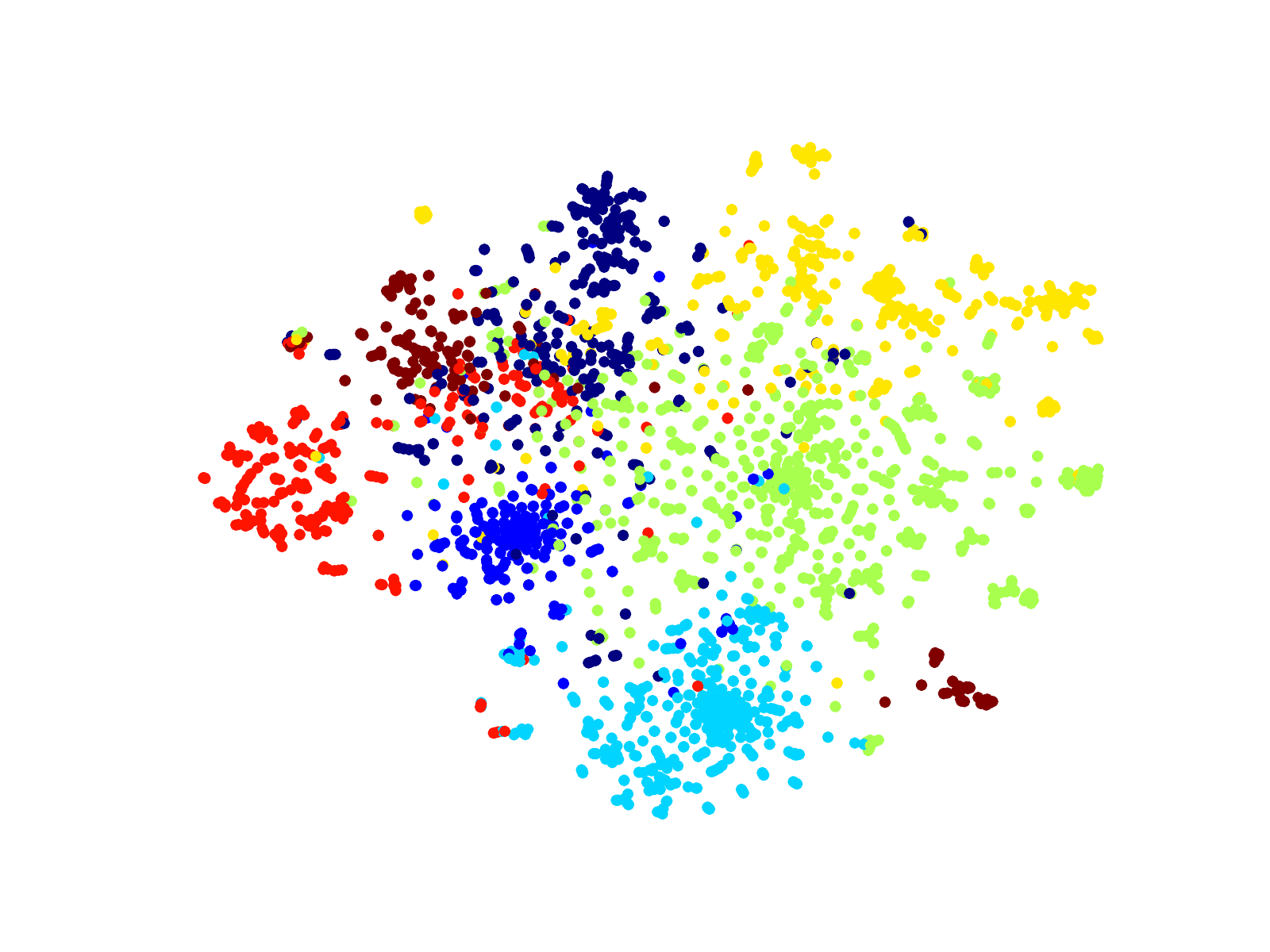}}\hspace{-1mm}
\subfigure[our 20]{
\includegraphics[width=0.12\textwidth]{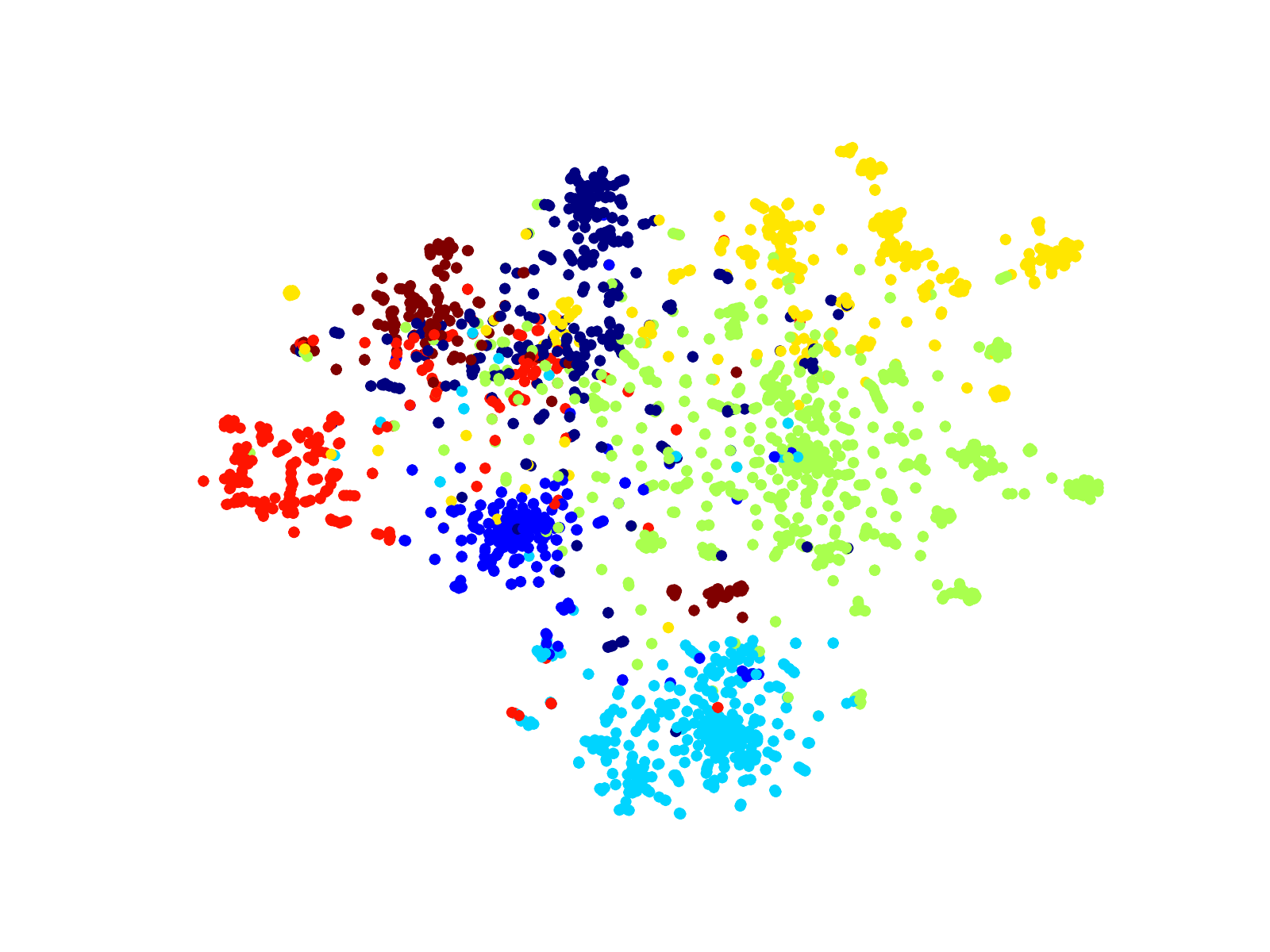}}\hspace{-1mm}
\subfigure[our 50]{
\includegraphics[width=0.12\textwidth]{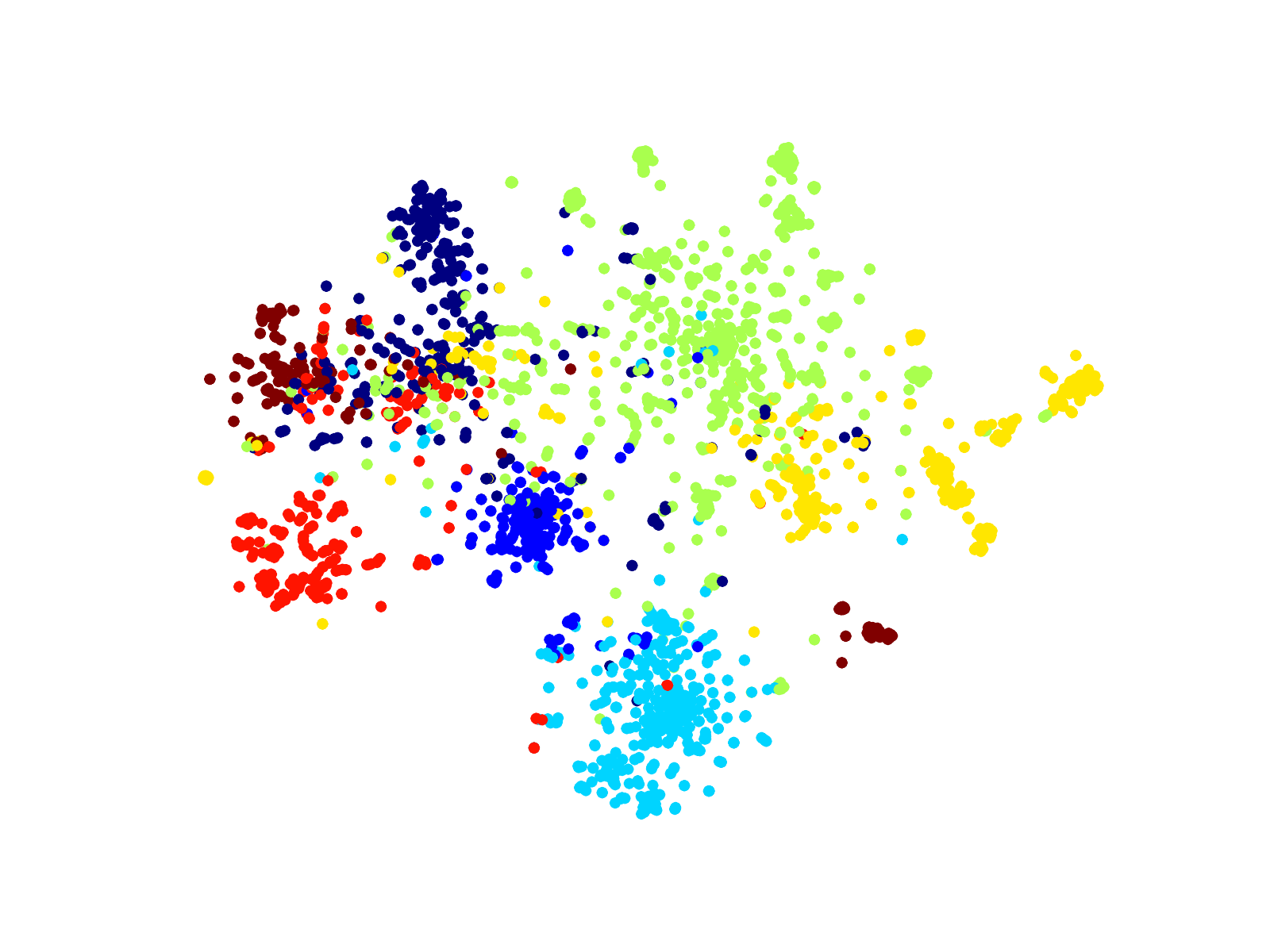}}\hspace{-1mm}
\subfigure[our 100]{
\includegraphics[width=0.12\textwidth]{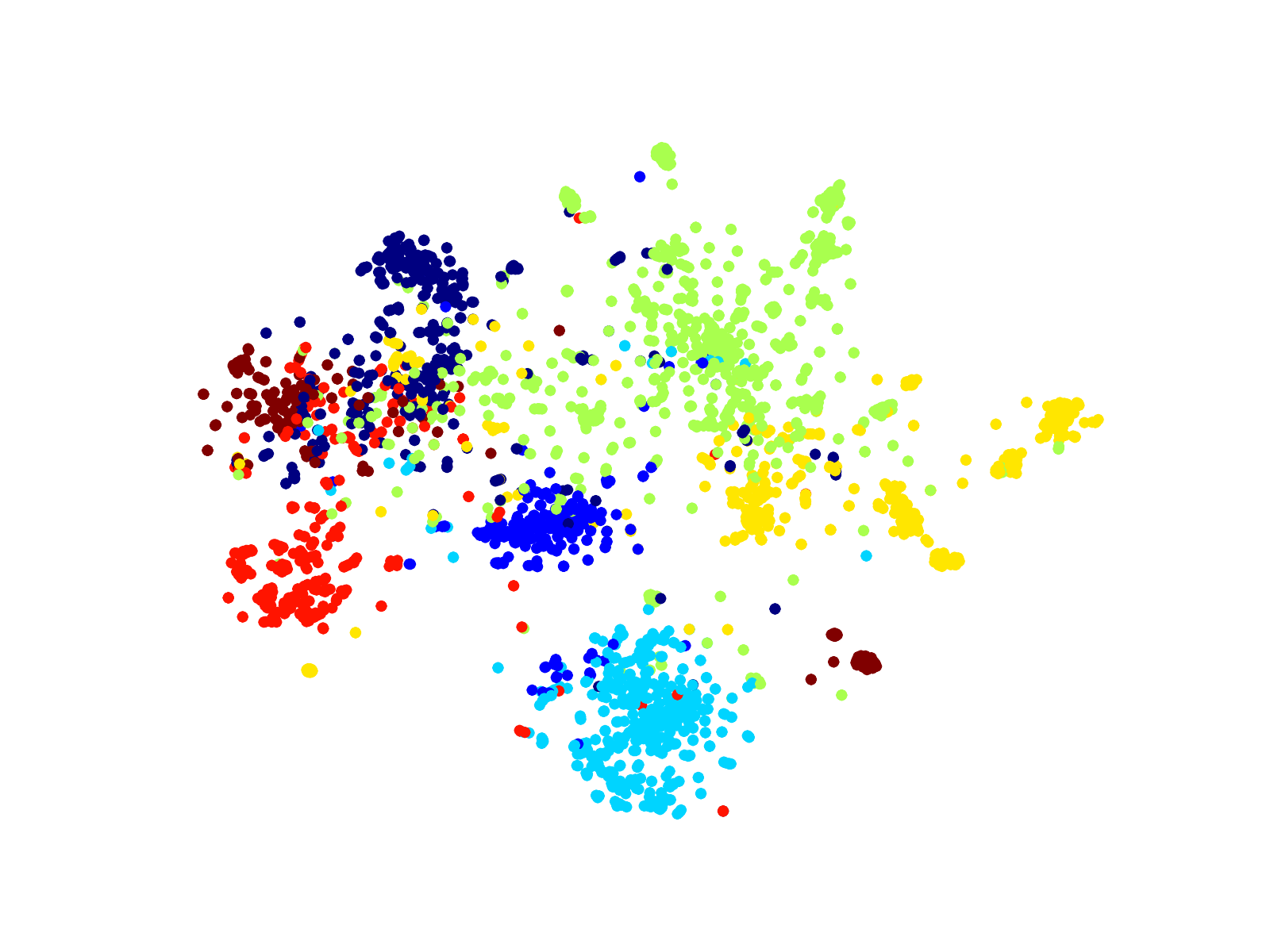}}\hspace{-1mm}
\subfigure[our 150]{
\includegraphics[width=0.12\textwidth]{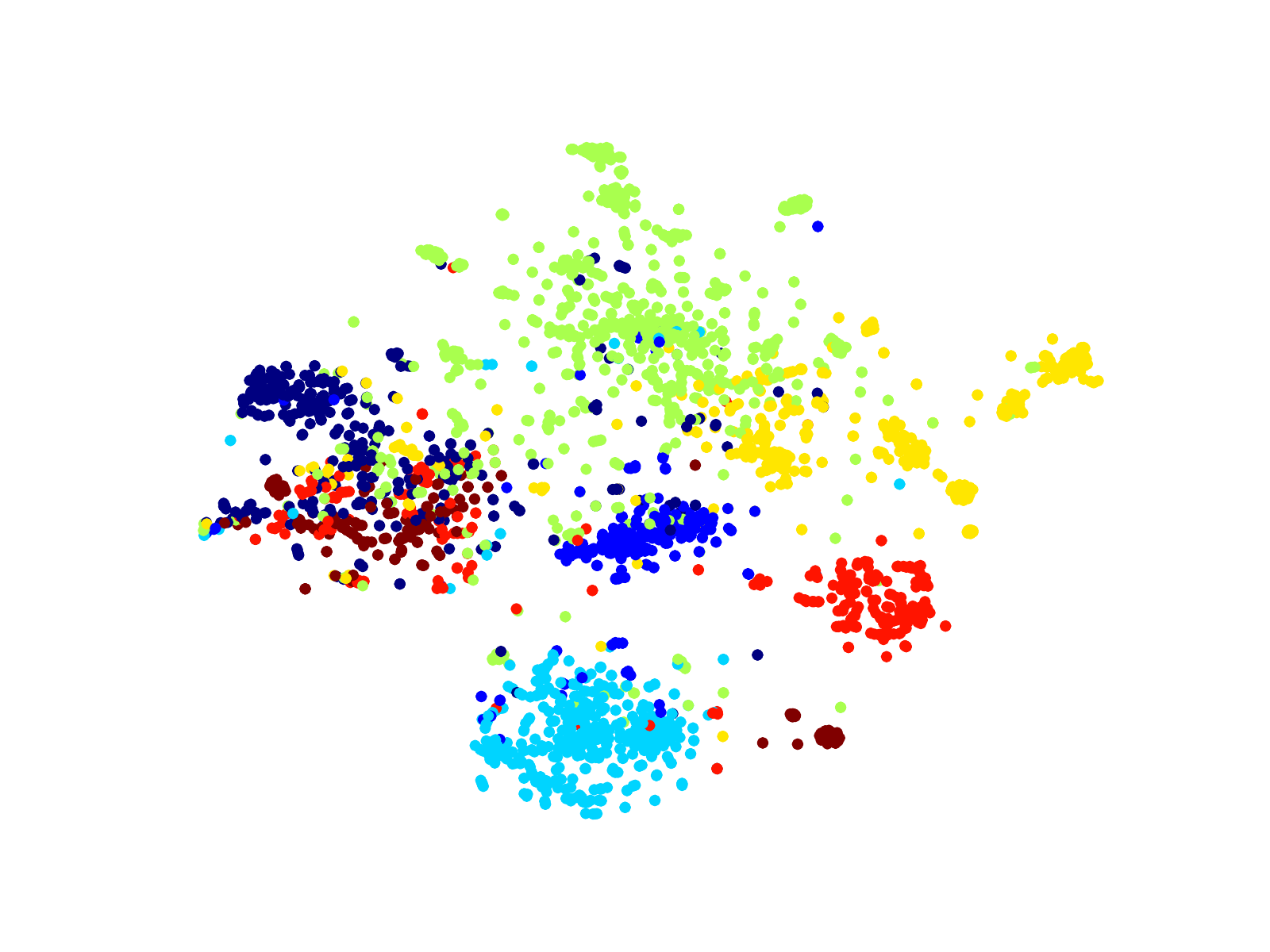}}\hspace{-1mm}
\subfigure[our 200]{
\includegraphics[width=0.12\textwidth]{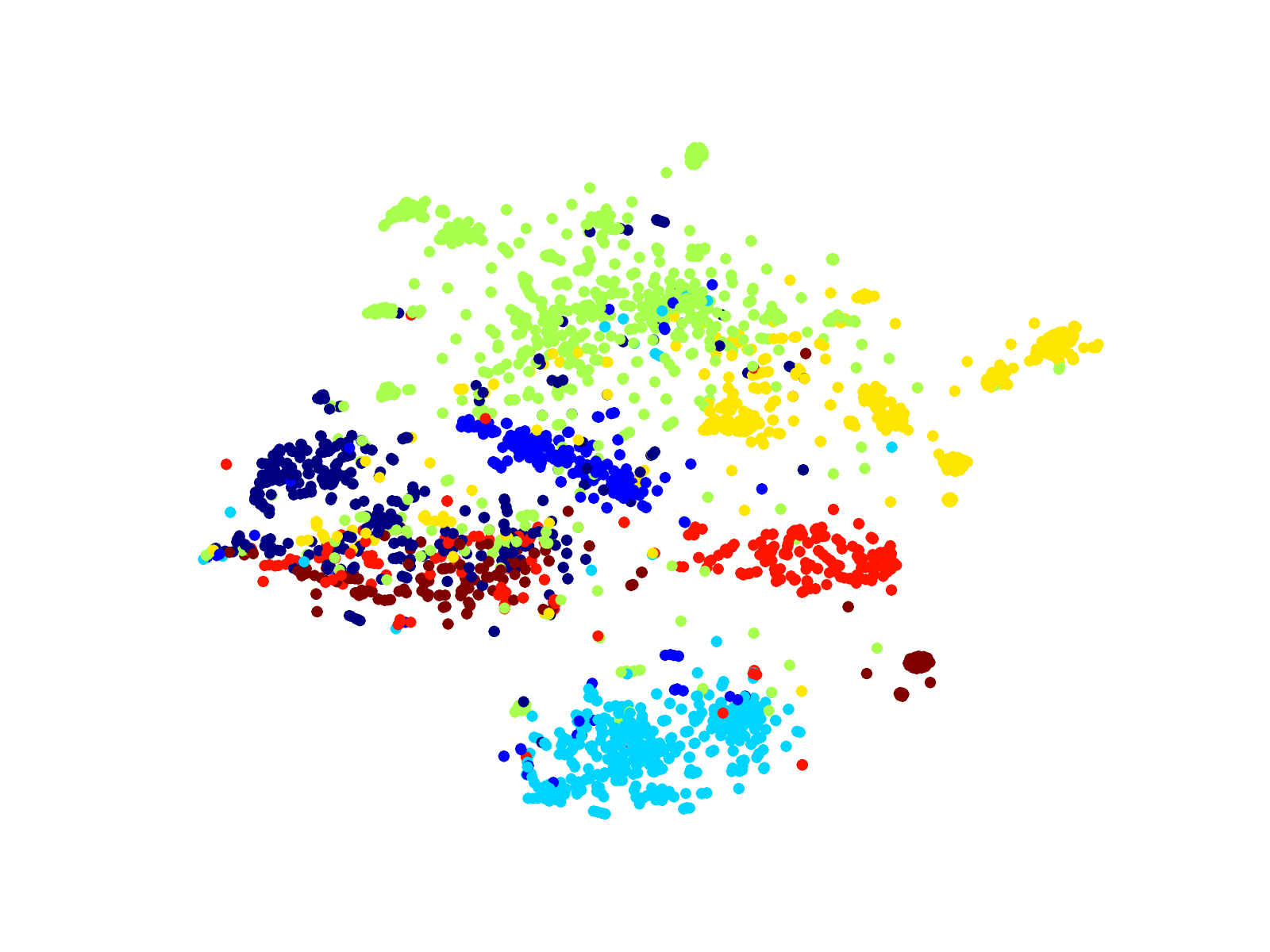}}\hspace{-1mm}

\subfigure[original]{
\includegraphics[width=0.12\textwidth]{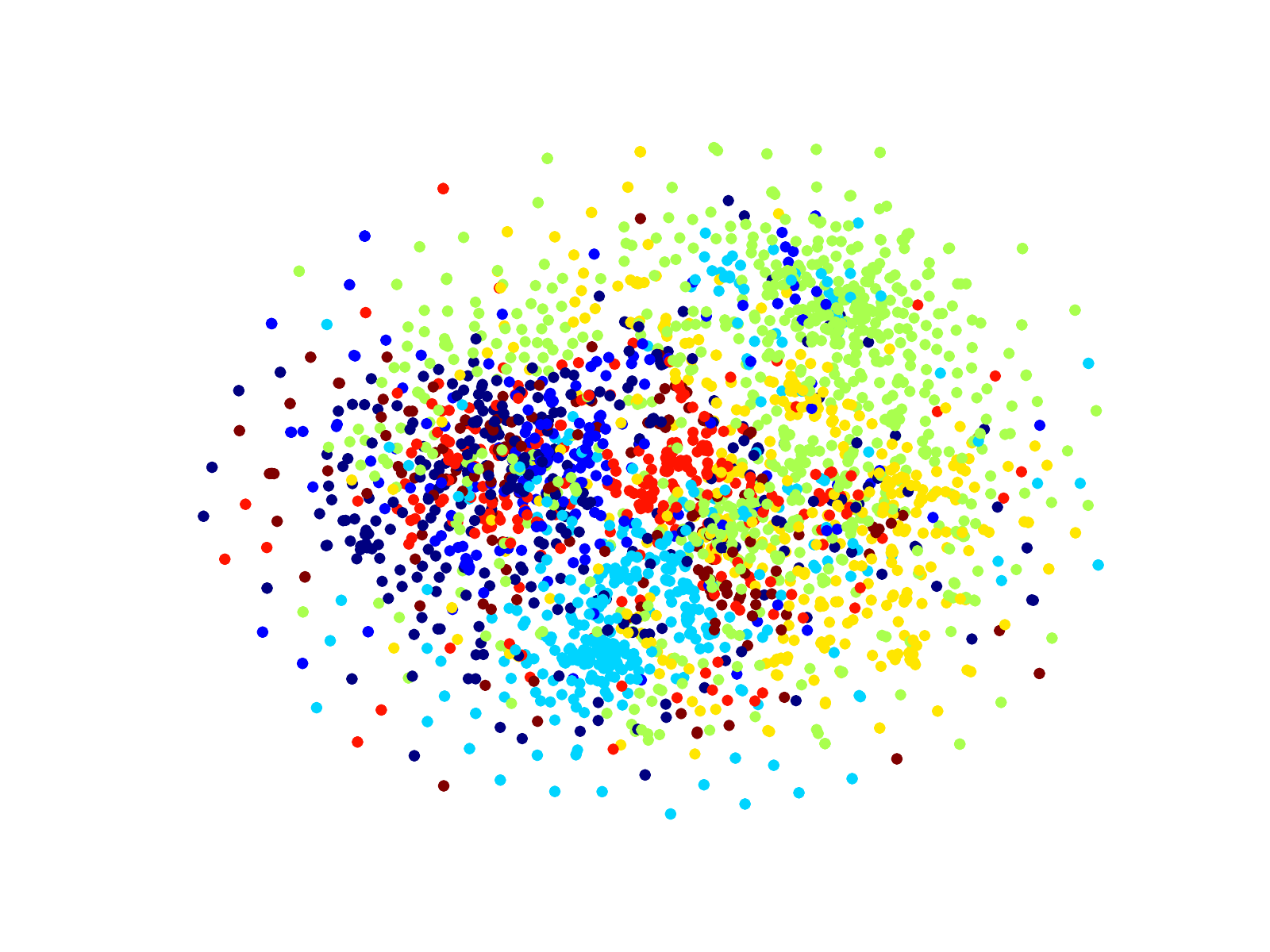}}\hspace{-1mm}
\subfigure[only 5]{
\includegraphics[width=0.12\textwidth]{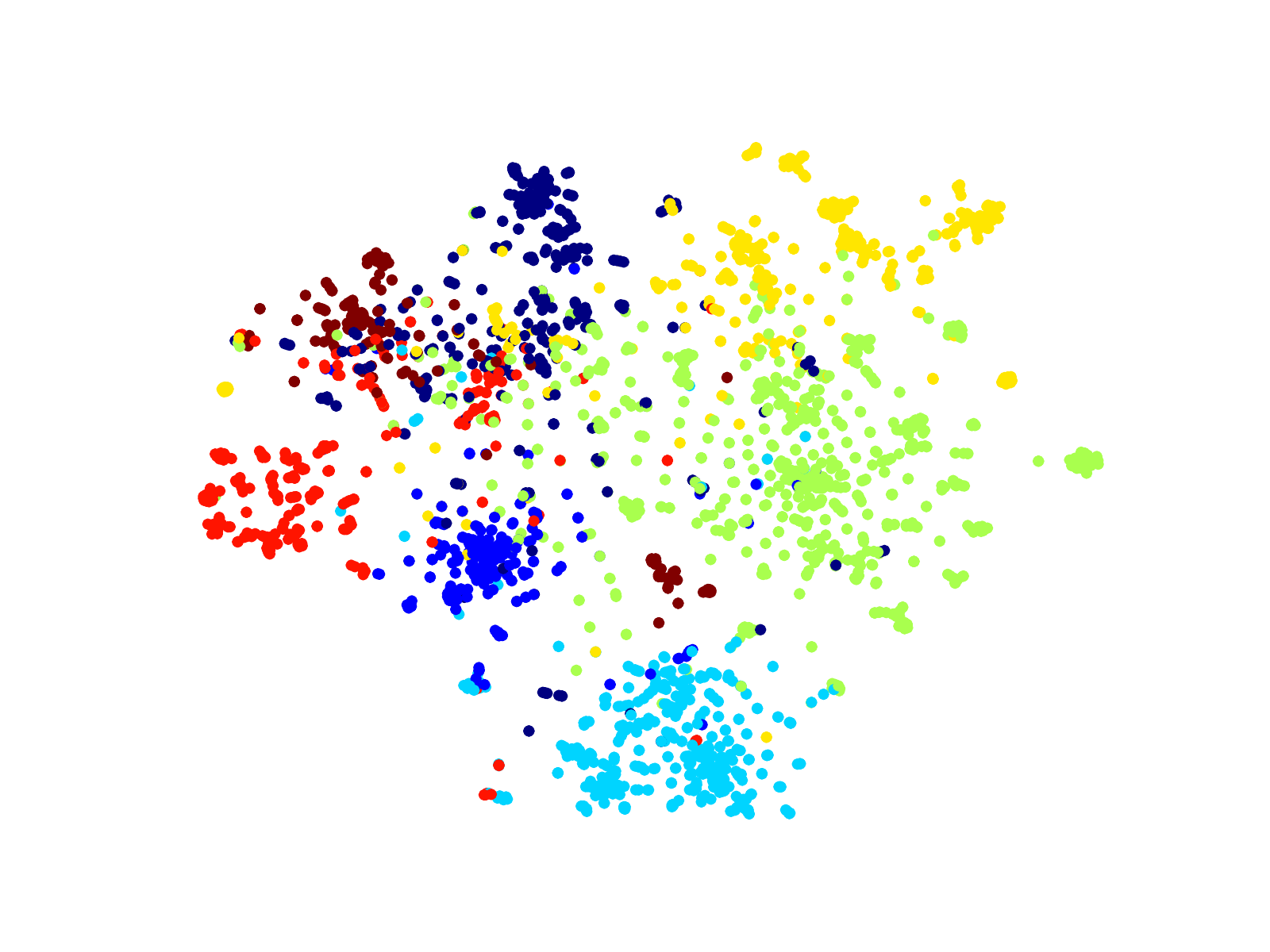}}\hspace{-1mm}
\subfigure[only 10]{
\label{sgc_hop_10}
\includegraphics[width=0.12\textwidth]{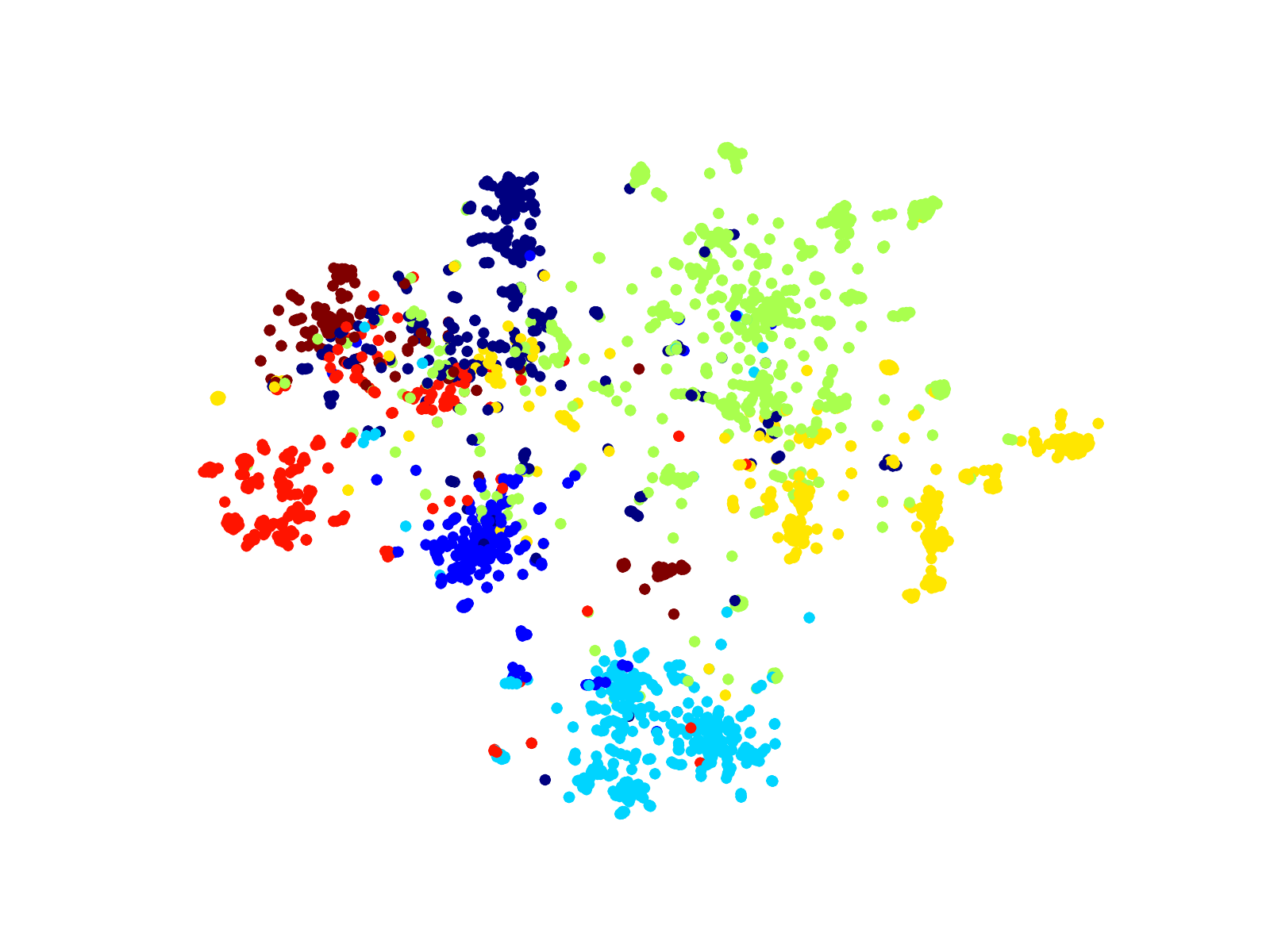}}\hspace{-1mm}
\subfigure[only 20]{
\includegraphics[width=0.12\textwidth]{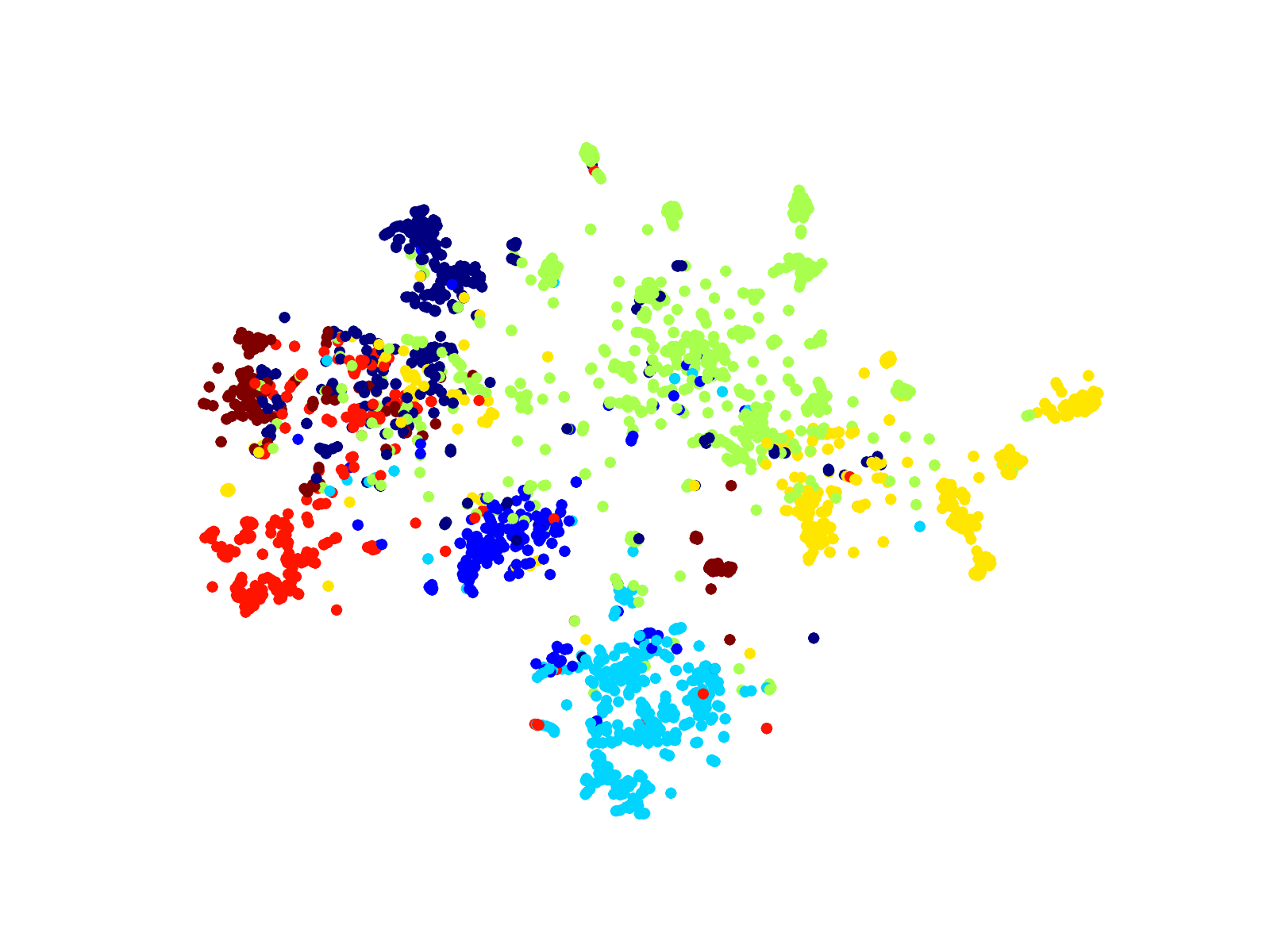}}\hspace{-1mm}
\subfigure[only 50]{
\includegraphics[width=0.12\textwidth]{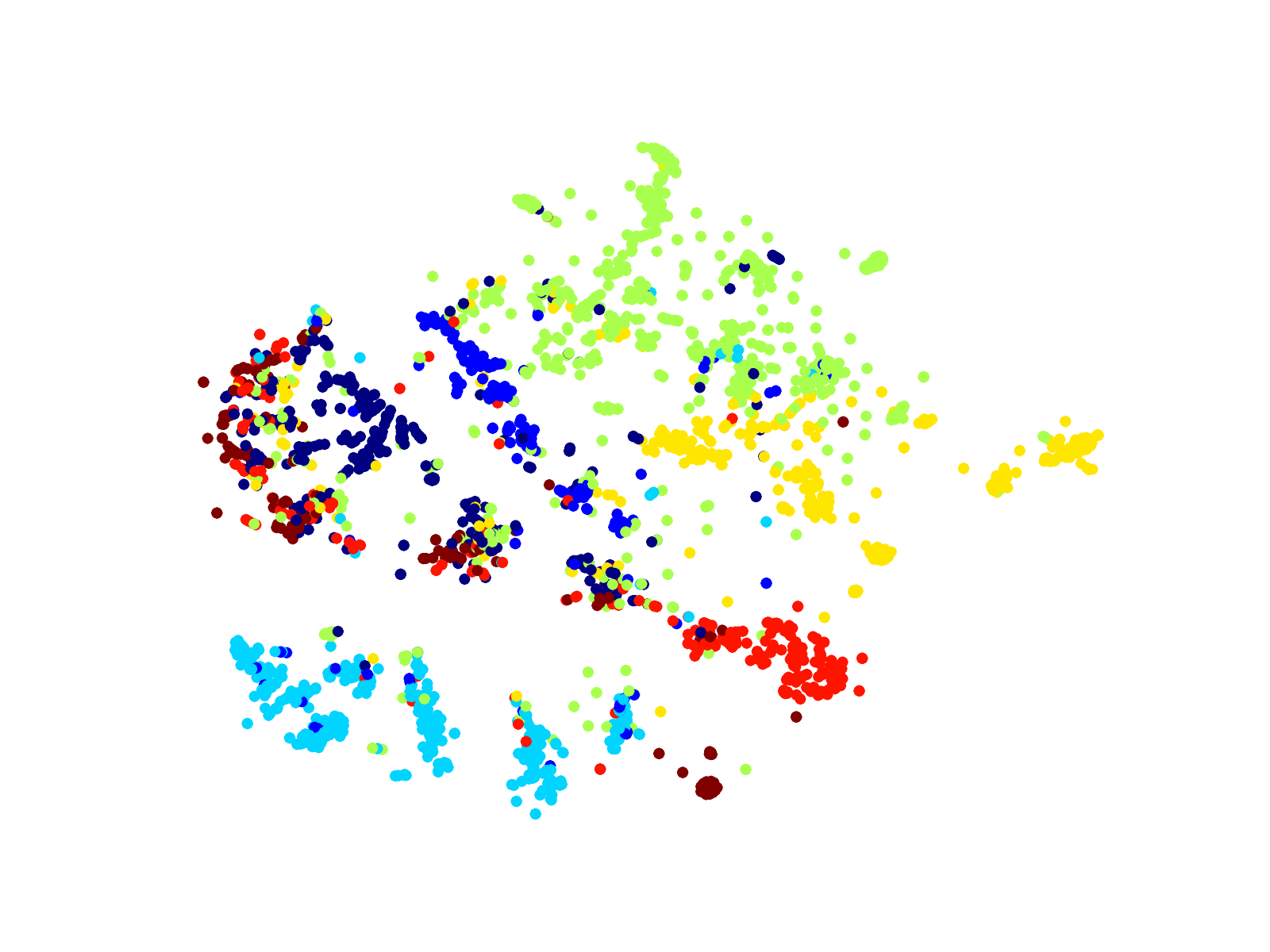}}\hspace{-1mm}
\subfigure[only 100]{
\label{sgc_hop_100}
\includegraphics[width=0.12\textwidth]{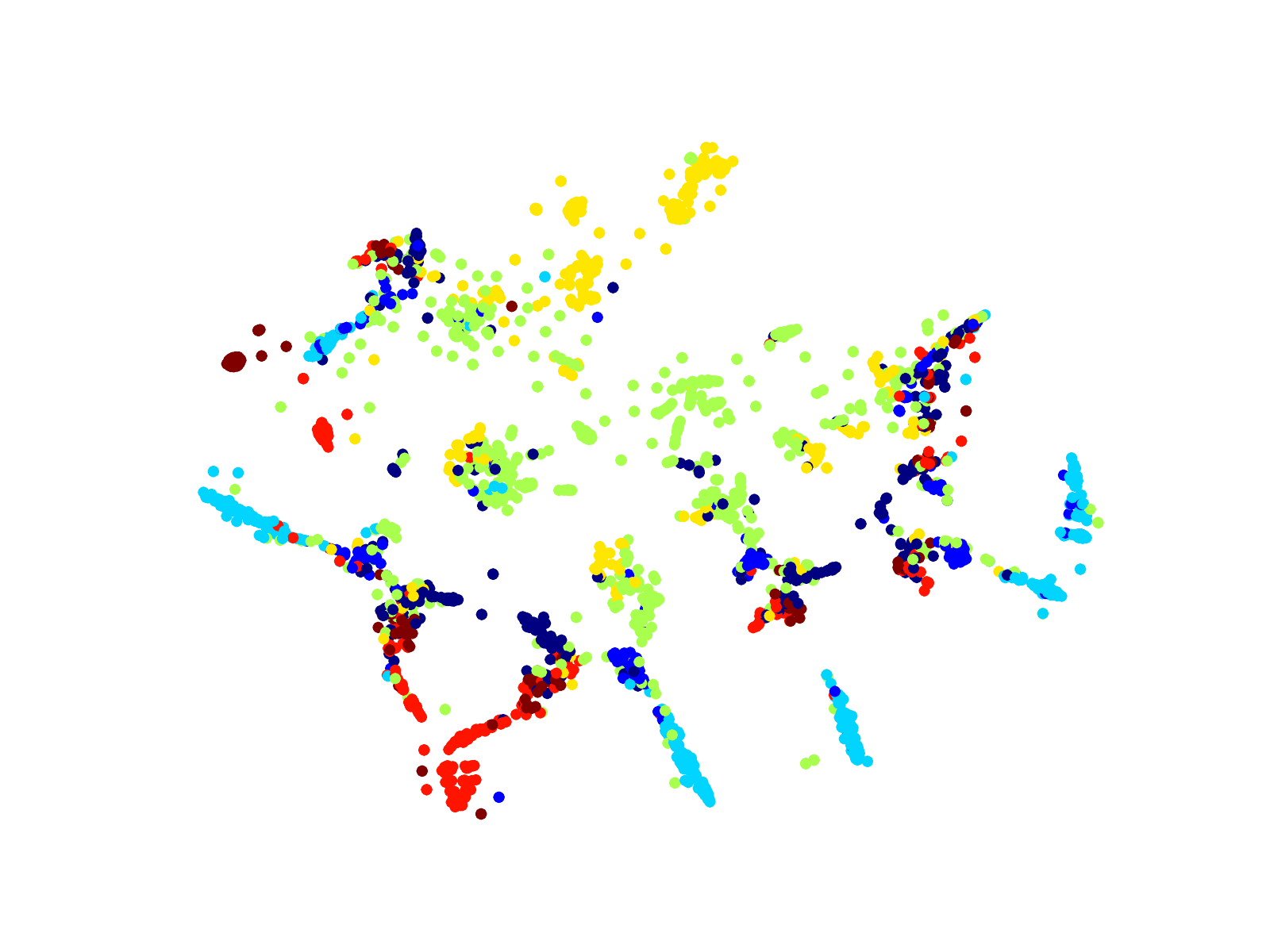}}\hspace{-1mm}
\subfigure[only 150]{
\label{sgc_hop_150}
\includegraphics[width=0.12\textwidth]{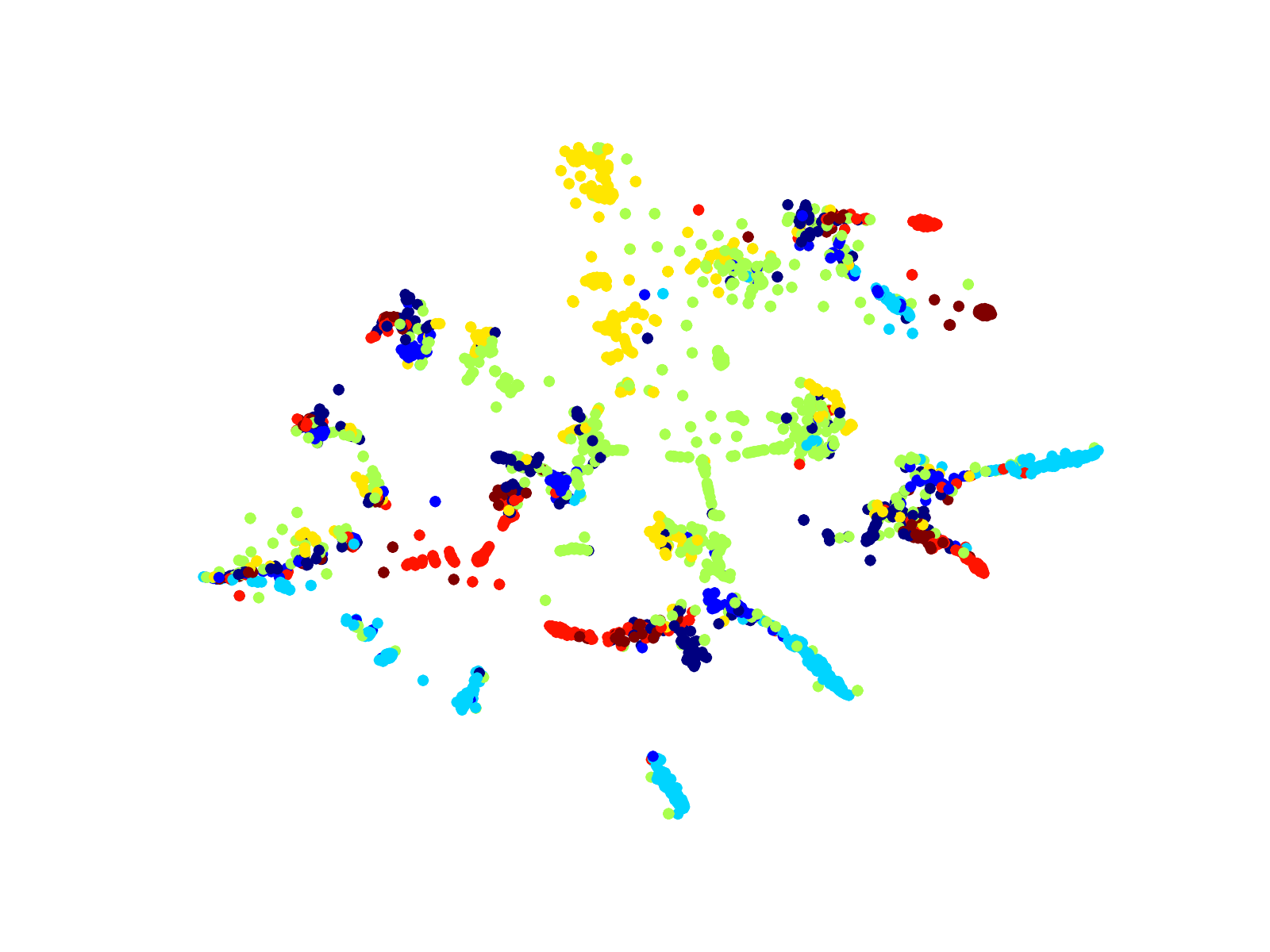}}\hspace{-1mm}
\subfigure[only 200]{
\label{sgc_hop_200}
\includegraphics[width=0.12\textwidth]{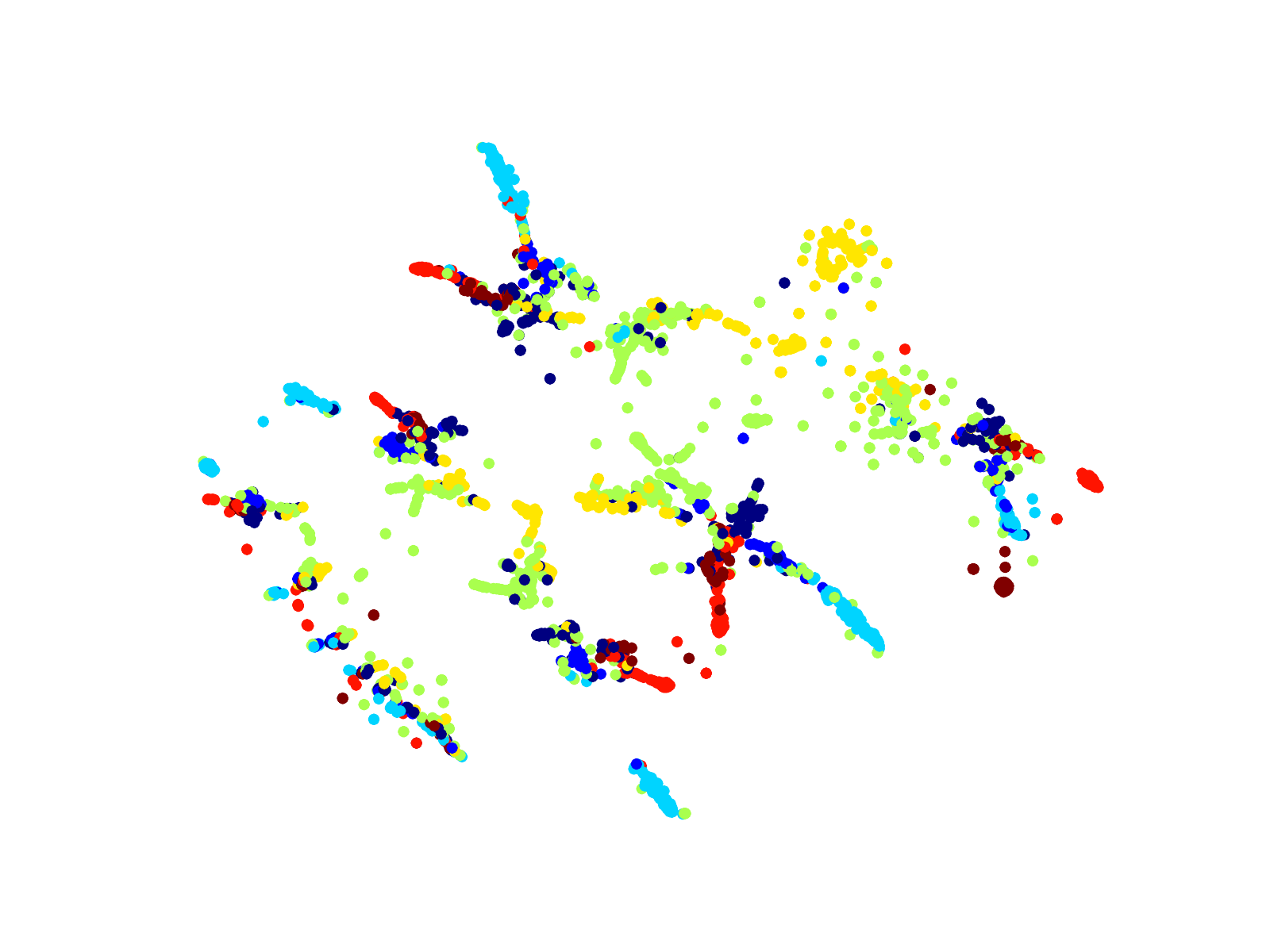}}\hspace{-1mm}
\caption{T-SNE visualization of node embedding on the Cora dataset.}
\label{fig.visualization}
\end{figure*}

% \begin{figure}[tbp]
%     \centering
%     \includegraphics[width=0.4\linewidth]{figure/efficiency.pdf}
%     \caption{Efficiency comparison on citation networks.}
%     \label{efficiency}
% \end{figure}

\subsection{Scalability Analysis}
A challenging task in graph representation learning is how to effectively get decent node embedding on very large graphs.
To verify the scalability advantage of NAFS, we add the evaluation on the large real-world datasets. The experiment settings are altered compared to Sec.~\ref{effi_compare}: both GAE and AGE are trained on an NVIDIA TITAN RTX, which has 24 GB of memory since only using CPU to train is unacceptable on large graphs in real-world applications.

Concretely, we sample subgraphs of different scales from the full ogbn-products graphs via uniformly random node selection and then report the sampled graph size (i.e., number of nodes) and the corresponding runtime (seconds) in Table~\ref{tab.scalability}. The experimental results show that NAFS can support the larger graphs (i.e., larger than 30,000 nodes) than the compared baselines. Besides, it is significantly faster than the compared baselines, especially for large graph datasets.
We also include more details in Appendix~\ref{scalability_comp} to demonstrate the excellent scalability of \sys (See Figure~\ref{fig.scalability}).

\begin{table}[tbp]
\centering
{
\noindent
\renewcommand{\multirowsetup}{\centering}
\caption{\small Scalablity comparison with different sample sizes on the ogbn-products dataset.}
\label{tab.scalability}
\resizebox{0.85\linewidth}{!}{
\begin{tabular}{cccccccc}
\toprule
\textbf{Methods}&\textbf{10,000}& \textbf{20,000}&\textbf{30,000}&\textbf{50,000}&\textbf{100,000}\\
\midrule
GAE & 10.2s & 36.3s & OOM & OOM & OOM\\
AGE & 68.5s & 256.2s& 727.7s&2315.7s&OOM\\
NAFS-mean & 2.8s  & 6.5s& 10.8s& 13.8s &23.7s\\
\bottomrule
\end{tabular}}}
% \vspace{-2mm}
\end{table}

\subsection{Ablation Study}
\label{ablation}
%\zwt{去掉weighted加权而是等权重加权，或者只选一个，3个任务3张图，纵轴在三个任务上的指标随着横轴深度的变化}
%\zwt{每张图三根曲线：1）去掉Ensemble，而是只用一个r 2）NAFS-Mean 3）NAFS-Max 4）NAFS-Concat 5）SIGN，等权重加起来，而且只用一个r}
%\zwt{1和5对比，说明我们基于α做feature engineering的好处。234和1对比，说明ensemble的作用}

To thoroughly investigate the proposed \sys, ablation studies on the node clustering task are designed to analyze the effectiveness of feature smoothing and feature ensemble in NAFS.
The experiments are conducted on the PubMed dataset, and Normalized Mutual Information (NMI) is used to measure the performance. 
% To interpret its effectiveness better, we also visualize the node embedding in Appendix~\ref{vis}.

\para{Different Weighting Strategies.}
One important operation in \sys is to use the node-adaptive weight to average the smoothed features of different smoothing steps.
Here we change the ``Adaptive Weight'' in our method to ``Single Hop'' (only the smoothed feature at the last smoothing step is reserved) and ``Naive Average'' (smoothed features at every smoothing step has equal weight) and evaluate their performance.
Figure~\ref{ablation_weight} shows the performance of these three different weighting strategies.
%During the experiment, the maximal aggregation depth ranges from 10 to 60, and the ``Mean'' ensemble strategy is adopted in NAFS.

From Figure~\ref{ablation_weight}, the weighting strategy ``Single Hop'' performs the worst among the three, since it only makes use of the information at the last smoothing step, which could lead to the over-smoothing issue when the maximal smoothing step becomes large.
On the other hand, the weighting strategy ``Adaptive Weight'' shows a better performance than ``Naive Average''.
Besides, when the number of maximal smoothing steps becomes large, the performance of ``Naive Average'' begins to drop, while ``Adaptive Weight'' does not.
``Naive Average'' assigns the same weight across all the different steps of smoothed features, which also leads to over-smoothing when it tries to exploit extremely deep structural information.
Instead, by assigning adaptive weights, the ``Adaptive Weight'' strategy in \sys could exploit such deep information and avoid the over-smoothing issue, thus improving the quality of the generated node embedding.

\para{Different Ensemble Strategies.} We use different ensemble strategies to obtain the final node embeddings in our proposed method - \sys, which include ``Mean'', ``Max'', and ``Concat''.
To evaluate the impacts of these different ensemble strategies, we change the maximal smoothing step, $K$, from 15 to 40, and evaluate corresponding node clustering performance.
For reference, the performance of NAFS without feature ensemble, ``r=0.3 only'' is also reported.
The experimental results in Figure~\ref{ablation_ensemble} illustrate that at most times, ``r=0.3 only'' performs worse than others, which shows the effectiveness of the three different ensemble strategies.
However, if we limit the comparison within the ensemble strategies, the performance superiority is unclear.
In Table~\ref{link_pre_performance} and~\ref{cluster_performance}, we can also have that different ensemble strategies perform diversely across different datasets and tasks. 
It indicates that these three ensemble strategies all have necessities in their own way.

\subsection{Interpretability}
\label{vis}
To better understand why \sys is effective, we visualize the node embedding generated by NAFS using T-SNE~\citep{van2008visualizing} on the Cora dataset.
Moreover, we also visualize the node embedding of $\mathbf{\hat{A}}^k\mathbf{X}$ for reference.
All the visualization results are shown in Figure~\ref{fig.visualization}.
The first row is the results of our proposed NAFS, and the second row is the results of $\mathbf{\hat{A}}^k\mathbf{X}$.

Figure~\ref{our_hop_10} and~\ref{sgc_hop_10} illustrate that at hop 10, the node embedding produced by NAFS and $\mathbf{\hat{A}}^k\mathbf{X}$ are both distinguishable.
But as the value of maximal smoothing step becomes larger, the node embedding of $\mathbf{\hat{A}}^k\mathbf{X}$ falls into total disorder, like the situation showed by Figure.~\ref{sgc_hop_100},~\ref{sgc_hop_150} and~\ref{sgc_hop_200}.
At the same time, NAFS is able to maintain the distinguishable results even when the value of the maximal number of smoothing step increases to 200.

\section{Conclusion}
This paper presents \sys, a novel graph representation learning method.
Unlike other GNN-based approaches, \sys focuses on improving the feature smoothing operation in the GNN layer and generating node embeddings in a training-free manner.
% Unlike other GNN-based approaches, \sys focuses on how to design a principled feature engineering approach to generate node embeddings in a training-free manner.
\sys proposes Node-Adaptive Feature Smoothing to generate smoothed features with adaptivity to each node's individual properties; it further employs feature ensemble to combine multiple smoothed features from diverse knowledge extractors effectively. 
Experiments results on typical tasks demonstrate that \sys performs comparably with or even outperforms the state-of-the-art GNNs, and at the same time enjoys high efficiency and scalability where NAFS shows its absolute superiority.

\section*{Acknowledgement}

This work is supported by NSFC (No. 61832001, 61972004), and PKU-Tencent Joint Research Lab. Bin Cui is the corresponding author.

\bibliography{reference}
\bibliographystyle{icml2022}

\clearpage

\appendix
\section{Theoretical Analysis}
\label{theory}
\subsection{What influences the Smoothing Weight?}
\label{t1}
The essential kernel of \sys is the Smoothing Weight, which determines the output results. We now analyze the factors affecting the value of Smoothing Weight. To simplify our analysis, we suppose $r=0$ in the normalized adjacency matrix and apply Euclidean distance as the distance function in Definition \ref{df1}. 
Thus we have
\begin{equation}
\small
D_{i}(k) = ||[\hat{\mathbf{A}}^{k}\mathbf{X}]_{i}-[\hat{\mathbf{A}}^{\infty}\mathbf{X}]_{i}||_{2},
\end{equation}
where $||\cdot||_{2}$ symbols two-norm.
\begin{theorem}
\label{thm1}
For any node $i$ in graph $\mathcal{G}$, there always exists
\begin{equation}
\small
D_{i}(k) \le \lambda_{2}^{k}\sqrt{\frac{\sum\limits_{j=1}\limits^{n}(\tilde{d_{j}}||\mathbf{X}_{j}||_{2}^{2})}{\tilde{d_{i}}}},
\end{equation}
where $\tilde{d_{i}}$ = $d_{i}+1$, $\tilde{d_{j}}$ = $d_{j}+1$, $||\mathbf{X}_{j}||_{2}$ denotes the two-norm of the initial feature of node $j$, and $0<\lambda_{2}<1$ denotes the second largest eigenvalue of the normalized adjacency matrix $\hat{\mathbf{A}}$. 
\end{theorem}

To prove Theorem \ref{thm1}, we introduce the following lemma.

\begin{lemma}
\small
\label{lemma2}
\begin{equation}
|(e_{i}\hat{\mathbf{A}}^{k})_{j}-(e_{i}\hat{\mathbf{A}}^{\infty})_{j}|\le \sqrt{\frac{\tilde{d_{j}}}{\tilde{d_{i}}}}\lambda_{2}^{k},
\end{equation}
where $e_{i}$ denotes a one-hot row vector with its $i^{th}$ components as 1 and other components as 0, $\lambda_{2}$ is the second largest eigenvalue of $\hat{\mathbf{A}}$ and $\tilde{d_{i}}$ denotes the degree of node $i$ plus 1 (to include itself).
\begin{equation*}
\small
\tilde{d_{i}} = d_{i}+1,\quad \tilde{d_{j}} = d_{j}+1,
\end{equation*}
\end{lemma}

The proof of Lemma \ref{lemma2} can be found in~\citep{chung1997spectral}. Next we will prove Theorem \ref{thm1}.
\begin{proof}
According to equation \ref{eq17}, we can have that
\begin{equation}
\small
\begin{split}
\small
D_{i}(k)
&= ||[\hat{\mathbf{A}}^{k}\mathbf{X}]_{i}-[\hat{\mathbf{A}}^{\infty}\mathbf{X}]_{i}||_{2}\\
&= ||(e_{i}\hat{\mathbf{A}}^{k}-e_{i}\hat{\mathbf{A}}^{\infty})\mathbf{X}||_{2}\\
&= \sqrt{\sum\limits_{j=1}\limits^{n}((e_{i}\hat{\mathbf{A}}^{k})_{j}-(e_{i}\hat{\mathbf{A}}^{\infty})_{j})^{2}\mathbf{X}_{j}^{2}}\\
&\le \sqrt{\lambda_{2}^{2k}\frac{\sum\limits_{j=1}\limits^{n}\tilde{d_{j}}\sum\limits_{p=1}\limits^{f}\mathbf{X}_{jp}^{2}}{\tilde{d_{i}}}}=\lambda_{2}^{k}\sqrt{\frac{\sum\limits_{j=1}\limits^{n}(\tilde{d_{j}}||\mathbf{X}_{j}||_{2}^{2})}{\tilde{d_{i}}}},\\
\end{split}
\end{equation}
where $\mathbf{X}_{jp}$ denotes the $p^{th}$ feature of node $j$.\\
\end{proof}
Based on Lemma \ref{lemma2} we then consider the smoothing distance for weighted averaged embedding features to the station state.
\begin{equation}
\small
    ||[\hat{\mathbf{A}}^{k}\mathbf{X}]_{i}-[\hat{\mathbf{A}}^{\infty}\mathbf{X}]_{i}||_{2}
\end{equation}
% \vspace{-2.5mm}
Therefore there holds Theorem \ref{thm1}
\begin{equation}
\small
D_{i}(k) \le \lambda_{2}^{k}\sqrt{\frac{\sum\limits_{j=1}\limits^{n}(\tilde{d_{j}}||\mathbf{X}_{j}||_{2}^{2})}{\tilde{d_{i}}}}.
\end{equation}
We denote the constant $\sum\limits_{j=1}\limits^{n}(\tilde{d_{j}}||\mathbf{X}_{j}||_{2}^{2})$ as $cdx$ because it is independent with $j$, then Theorem \ref{thm1} can be written as
\begin{equation}
\small
\label{eq20}
D_{i}(k) \le \lambda_{2}^{k}\sqrt{\frac{cdx}{\tilde{d_{i}}}}.
\end{equation}
We then analyze the factors affecting the Smoothing Weight on a specific node $v_{i}$.
From Eq. \ref{eq20} we know that the nodes with smaller degrees may have larger $D_{i}(k)$. Combined with definition \ref{df3}, we infer that larger $D_{i}(k)$ makes the $\max_{k}D_{i}(k)$ more dominant after the softmax operation, causing that weighted average results depend more on itself and its near neighbors. 
Inversely, for the nodes with smaller degrees, its result of weighted average depends more equally on all itself, its near neighbors, and its distant neighbors.

At the same time, the Smoothing Weight of the node in a sparser graph ($\lambda_{2}$ is positively relative with the sparsity of a graph) decays slower as k increases. Thus, the weighted average result depends more equally on itself, its near neighbors and its distant neighbors. While for the nodes in a denser graph, the weighted average result depends more on its near neighbors and itself.

\subsection{How \sys prevent over-smoothing?}
\label{t2}
Based on Theorem \ref{thm1} we then consider the smoothing distance for weighted averaged embedding features to the station state:
\begin{equation}
\small
\begin{aligned}
&||\sum_{k=0}^{K}\omega_i(k)*[\hat{\mathbf{A}}^{k}\mathbf{X}]_{i}-[\hat{\mathbf{A}}^{\infty}\mathbf{X}]_{i}||_{2}\\
=&||\sum_{k=0}^{K}\omega_i(k)*([\hat{\mathbf{A}}^{k}\mathbf{X}]_{i}-[\hat{\mathbf{A}}^{\infty}\mathbf{X}]_{i})||_{2}\\
\le &\sum_{k=0}^{K}\omega_i(k)*||[\hat{\mathbf{A}}^{k}\mathbf{X}]_{i}-[\hat{\mathbf{A}}^{\infty}\mathbf{X}]_{i}||_{2}\\
=&\sum_{k=0}^{K}\omega_i(k)*D_i(k).
\end{aligned}
\end{equation}
When $\omega_i(k) = 1/(K+1)$, which means the average option is non-weighted, we have
\begin{equation}
\begin{aligned}
\small
&\lim_{K\to \infty} \sum_{k=0}^{K}\omega_i(k)*D_i(k)\\
=&\lim_{K\to \infty} \frac{1}{K+1}\sum_{k=0}^{K}D_i(k)\\
\le & \lim_{K\to \infty} \frac{1}{K+1}\sum_{k=0}^{K}\lambda_{2}^{k}\sqrt{\frac{cdx}{\tilde{d}_i}}\\
=&\lim_{K\to \infty} \frac{1}{K+1}\frac{1-\lambda_{2}^{K+1}}{1-\lambda_{2}}\sqrt{\frac{cdx}{\tilde{d}_i}}\\
=&0,
\end{aligned}
\end{equation}
causing over-smoothing.

When $\omega_i(k) = D_i(k)/(\sum_{l=0}^{K}D_i(l))$, let $\mathbf{X}_i^k = [\hat{\mathbf{A}}^{k}\mathbf{X}]_{i}-[\hat{\mathbf{A}}^{\infty}\mathbf{X}]_{i}$, we have: 
\begin{equation}
\begin{split}
\small
||\hat{\mathbf{X}}_i-[\hat{\mathbf{A}}^{\infty}\mathbf{X}]_i||_{2}
&=\lim_{K\to \infty}\omega_i(0)||\mathbf{X}_i^0+\sum_{k=1}^{K}\frac{\omega_i(k)}{\omega_i(0)}\mathbf{X}_i^k||_{2}.
\end{split}
\end{equation}

In real-world graphs, nodes have different initial features, thus there is little chance that the combination of a node's neighboring features both lies in the opposite direction and is of the same norm of the node's initial feature.
Suppose that there exits a constant $\epsilon > 0$ satisfying $\min\left(D_i(0),||\mathbf{X}_i^0+\sum_{k=1}^{K}\frac{\omega_i(k)}{\omega_i(0)}\mathbf{X}_i^k||_{2}\right) \geq \epsilon$ for node $i$,
we have:
\begin{equation}
\small
\begin{split}
||\hat{\mathbf{X}}_i-[\hat{\mathbf{A}}^{\infty}\mathbf{X}]_i||_{2} & \geq  \frac{\epsilon^2}{\lim_{K\to \infty}\sum_{k=0}^{K}D_i(k)} \\
&\ge \frac{\epsilon^2}{\frac{1}{1-\lambda_{2}}\sqrt{\frac{cdx}{\tilde{d}_i}}} > 0
\end{split}.
\end{equation}

Thus we see that even $K$ goes to infinity, we are able to prevent the node representations from reaching the stationary state (the distance bound depends on the node degree $d_i$ and the initial feature $\mathbf{X}_i$). Note that in practice the representation at $k$-th smoothing step $\mathbf{X}^{(k)}$ achieves the stationary state much earlier than the infinity-th smoothing step we use in the theoretical analysis, so we use the softmax normalization in Equation~\ref{iw} to produce a slightly larger bias towards features with longer distances to the stationary state.

\subsection{The node-adaptive weighting in \sys.}
\label{t3}
The above analysis \emph{theoretically proved} our weighting scheme is able to prevent
over-smoothing as the smoothing step goes to infinity. We now show that another advantage of our method is that it can fully leverage the multiple features over different smooth steps in a node-adaptive manner, which is different from the traditional routine of the smooth-blind and fixed scheme.
Suppose that the feature of $k$ step is not over-smoothed yet for a specific node $i$, i.e., the distance of $D_i(k) = \epsilon_i>0$,
we have:
\begin{equation}
\small
\begin{split}
\omega_i(k) =  \frac{\epsilon_i}{\lim_{K\to \infty}\sum_{k=0}^{K}D_i(k)} \ge \frac{\epsilon_i}{\frac{1}{1-\lambda_{2}}\sqrt{\frac{cdx}{\tilde{d}_i}}} > 0
\end{split}.
\end{equation}
We see that as long as the feature is not over-smoothed, our method will assign a non-zero weight to the feature. Further, we see that the bound of weight can be affected by the over-smoothing distance of $\epsilon_i$ and degree $d_i$ of a specific node, implying an adaptive weighting strategy.

% we suppose $D_i(0) \ge 0$,
% \begin{equation}
% \begin{split}
% \lim_{K\to \infty} \sum_{k=0}^{K}D_i(k)
% &\le\lim_{K\to \infty} \frac{1+\lambda_{2}^{K}}{1-\lambda_{2}}\sqrt{\frac{cdx}{\tilde{d}_i}}\\
% &=\frac{1}{1-\lambda_{2}}\sqrt{\frac{cdx}{\tilde{d}_i}}
% \end{split},
% \end{equation}
% and thus 
% \begin{equation}
% \omega_i(k) \ge D_i(k)({1-\lambda_{2}})\sqrt{\frac{\tilde{d}_i}{cdx}}.
% \end{equation}
% Therefore the sign of $w_i(k)$ is determined by the sign of $D_i(k)$, which is not affected by the maximum smoothing step $K$.
% When $K$ goes to infinity, if the stationary state is not reached, the absolute value of $w_i(k)$ is not zero, which preserves the personalized information of each node.
% Thus, this weighting mechanism can alleviate the over-smoothing issue compared with assigning identical weights.
% \begin{equation}
% \begin{split}
% \lim_{K\to \infty} \sum_{k=0}^{K}\omega_i(k)*D_i(k)
% &\ge \omega_i(0)*D_i(0) \\
% &\ge D_i(0)^2({1-\lambda_{2}})\sqrt{\frac{\tilde{d}_i}{cdx}},\\
% &\ge 0
% \end{split}
% \end{equation}
% which avoid over-smoothing.

%\zwt{怎么引出softmax}

\section{Details on the Experiments}
\subsection{Datasets Description}
\label{data}
Cora, Citeseer, PubMed, and Wiki are four popular network datasets.
The first three~\citep{yang2016revisiting} are citation networks where nodes stand for research papers, and an edge exists between a node pair if one cites the other.
Wiki~\citep{yang2015network} is a webpage network where nodes stand for webpages, and an edge exists between a node pair if one links the other.
The ogbn-arxiv~\citep{hu2020ogb} dataset is also a citation network that contains more than 160M nodes. Besides, the ogbn-products dataset is an undirected and unweighted graph, representing an Amazon product co-purchasing network.
% an undirected and unweighted graph, representing an Amazon product co-purchasing network. 
%In Cora and Citeseer, the node features are binary word vectors; while in PubMed and Wiki, each node has a TF-IDF weighted word vector.
Table~\ref{Dataset} presents an overview of these six datasets.

\subsection{Hyperparameters setting}
\label{HPO}
When generating node embeddings, we use the values of $r$ in [0, 0.1, 0.2, 0.3, 0.4, 0.5] to get six different normalized adjacency matrix $\mathbf{\hat{A}}$.
The only exception is the link prediction task on the PubMed dataset, where we use values of $r$ in [0.3, 0.4, 0.5] instead.
%Furthermore, these results are combined using different ensemble strategies.
The optimal value of the maximal smoothing steps ranges from 1 to 70.
%The mean values and the variances of different performance metrics are reported.
Hyperparameters for all the baseline methods are tuned with OpenBox~\citep{li2021openbox} or following the settings in their original paper.

\begin{table}[tbp]
\centering
{
\noindent
\renewcommand{\multirowsetup}{\centering}
\caption{\small Overview of the datasets.}
\label{Dataset}
\resizebox{0.8\linewidth}{!}{
\begin{tabular}{ccccccc}
\toprule
\textbf{Dataset}&\textbf{Nodes}& \textbf{\#Features}&\textbf{\#Edges}&\textbf{\#Classes}\\
\midrule
Cora & 2,708 & 1,433 & 5,429 & 7 \\
Citeseer & 3,327 & 3,703 & 4,732 & 6 \\
PubMed & 19,717  & 500 & 44,338 & 3\\
Wiki & 2,405 & 4,973 & 17,981 & 17 \\
%ACM & 3,025 & 1,870 & - & 3\\
%DBLP & 4,058 & 334 & - & 4\\
ogbn-arxiv & 169,343	& 128 & 1,166,243 & 40\\
ogbn-products & 2,449,029  & 100 & 61,859,140 & 47\\
%ogbn-papers100M & 111,059,956 & 100 & 1,615,685,872 & 47\\
\bottomrule
\end{tabular}}}
\end{table}

\subsection{Experimental Environment}
\label{envi}
The experiments are conducted on a machine with Intel(R) Xeon(R) Gold 5120 CPU @ 2.20GHz, and a single NVIDIA TITAN RTX GPU with 24GB GPU memory.
The operating system of the machine is Ubuntu 16.04.
For software versions, we use Python 3.6, Pytorch 1.7.1, and CUDA 10.1.

\begin{figure*}[tp!]
\centering  
\subfigure{
\label{scala_effi}
\includegraphics[width=0.3\textwidth]{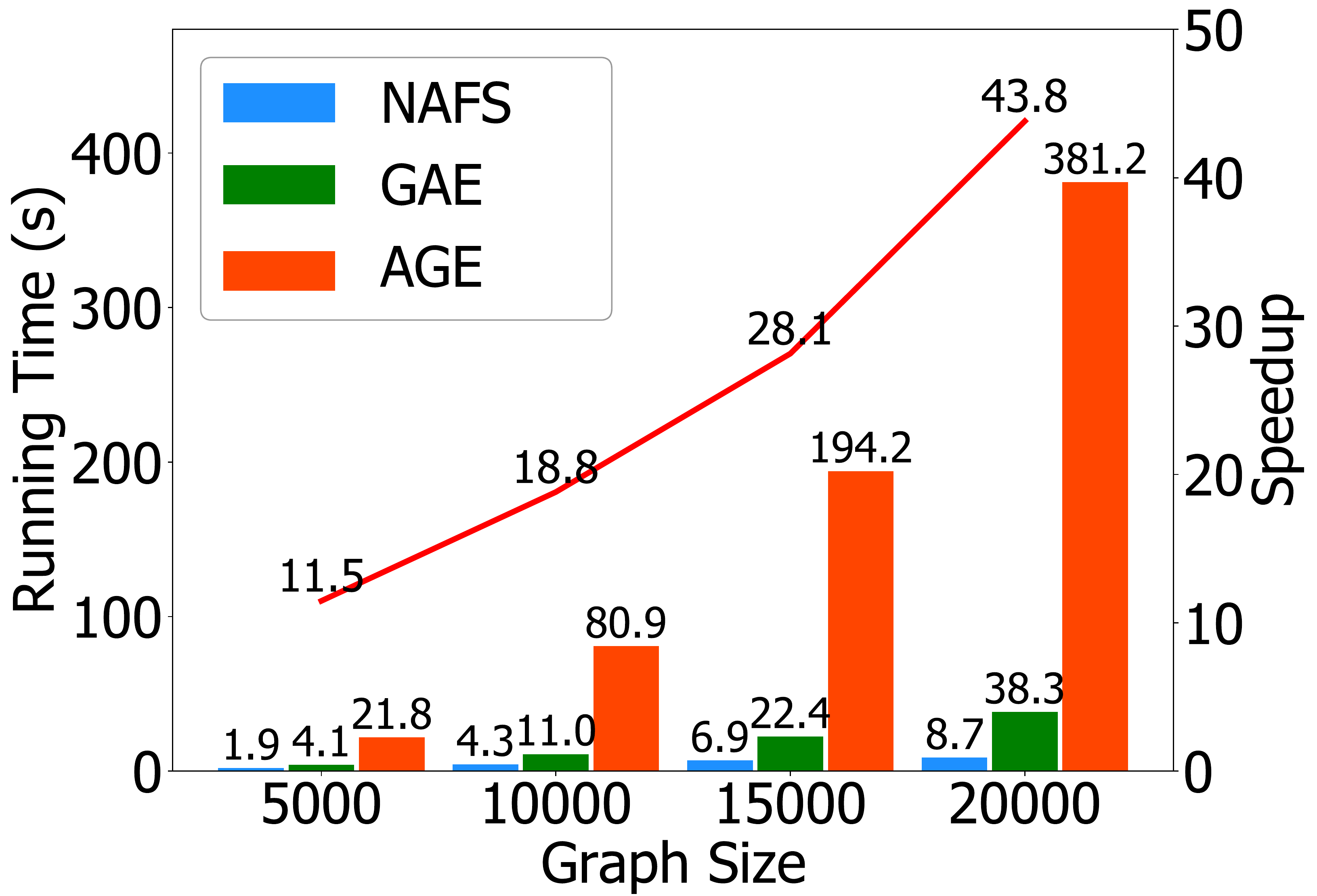}}\hspace{3mm}
\subfigure{
\label{scala_scala}
\includegraphics[width=0.3\textwidth]{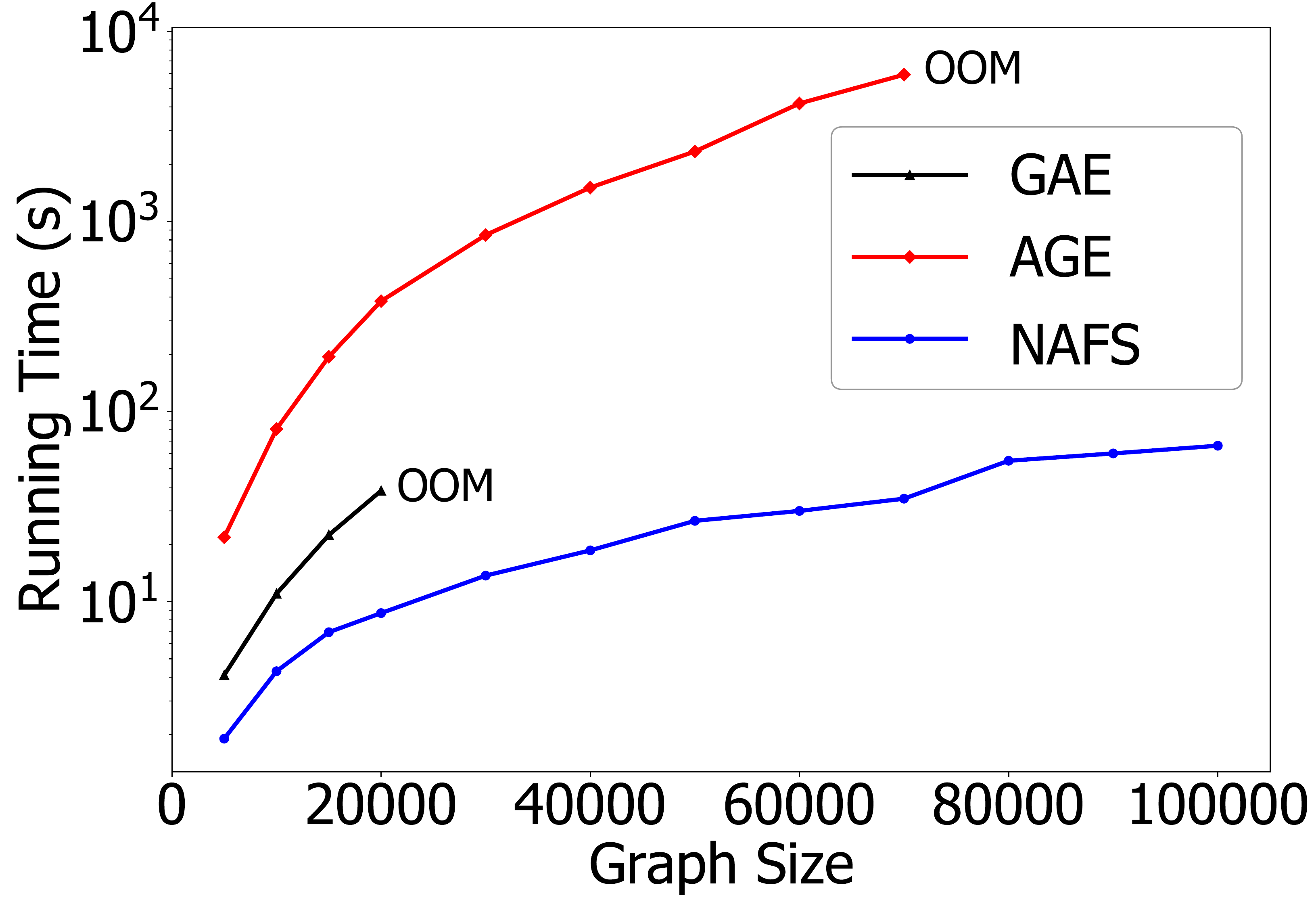}}
\caption{Scalablity comparison on synthetic graphs, OOM is ``out of memory''.}
\label{fig.scalability}
\end{figure*}

\section{Additional empirical results}

\subsection{Performance on the Node Classification Task}
\para{Node classification performance.} In this part, we assess the quality of the node embeddings generated by NAFS by the evaluation on the node classification task.
We follow the linear evaluation protocol, which applies a linear classifier (i.e., Logistic Regression) to the node embeddings to generate final prediction results.
We choose GCN~\citep{kipf2016semi}, JK-Net~\citep{xu2018representation}, C\&S~\citep{huang2020combining}, SGC~\citep{wu2019simplifying}, GAT~\citep{velivckovic2017graph}, PPRGo~\citep{bojchevski2020scaling}, APPNP~\citep{klicpera2018predict}, and DAGNN~\citep{liu2020towards} as comparison baselines on the node classification task.
Note that C\&S in our evaluation adopts a two-layer MLP as the base model.
The evaluation results on three popular datasets, Cora, Citeseer, and PubMed~\citep{yang2016revisiting}, are provided in Table~\ref{perf_classi}.

The table shows that our method achieves comparable predictive accuracy as one of the state-of-the-art methods DAGNN. 
However, since the node embeddings are generated before training, our method avoids performing recursive feature smoothing at each training epoch and storing the entire adjacency matrix on GPU. 
In this way, our method is more scalable and efficient to apply on large graphs than most state-of-the-art methods like DAGNN.

\begin{table}[tpb!]
%\vspace{-5mm}
\caption{Test accuracy on the node classification task.}
%\vspace{-2mm}
\centering
{
\noindent
\renewcommand{\multirowsetup}{\centering}
\resizebox{0.8\linewidth}{!}{
\begin{tabular}{c|ccc}
\toprule
\textbf{Methods} & \textbf{Cora} & \textbf{Citeseer} & \textbf{PubMed} \\
\midrule
GCN & 81.8$\pm$0.5 & 70.8$\pm$0.5 &79.3$\pm$0.6 \\
JK-Net & 81.9$\pm$0.4  & 70.7$\pm$0.7 & 78.8$\pm$0.7 \\
C\&S & 76.7$\pm$0.4 & 70.8$\pm$0.6 & 76.5$\pm$0.5 \\
SGC & 81.0$\pm$0.2 & 71.3$\pm$0.5 &78.9$\pm$0.5 \\
GAT & 83.0$\pm$0.7 & 72.5$\pm$0.6 &79.0$\pm$0.3 \\
PPRGo & 82.4$\pm$0.3 & 71.3$\pm$0.5 &80.0$\pm$0.4 \\
APPNP & 83.3$\pm$0.5 & 71.8$\pm$0.5 & 79.7$\pm$0.3\\
DAGNN & \textbf{84.4$\pm$0.6} & 73.3$\pm$0.6 & \textbf{80.5$\pm$0.5}\\
\midrule 
NAFS-mean & \underline{84.2$\pm$0.6} & 73.2$\pm$0.5 & \underline{80.5$\pm$0.6}\\
NAFS-max & 83.8$\pm$0.7 & \underline{73.4$\pm$0.6} & 80.4$\pm$0.6\\
NAFS-concat & 84.1$\pm$0.6 & \textbf{73.5$\pm$0.4} & 80.3$\pm$0.4\\
\bottomrule
\end{tabular}}}
\label{perf_classi}
\end{table}

\begin{figure}
% \vspace{-5mm}
	\centering
	\includegraphics[width=.8\linewidth]{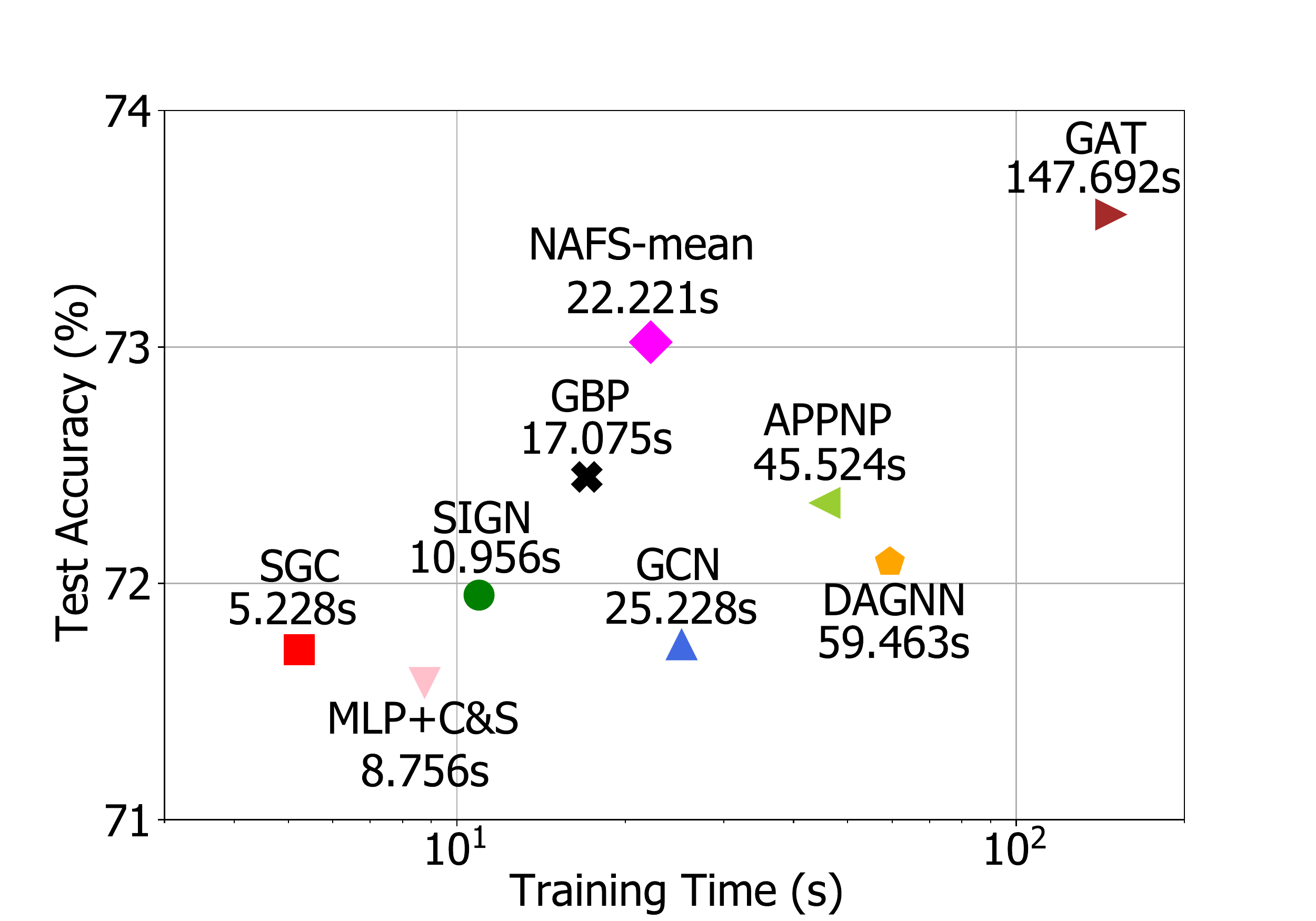}
 	% \vspace{-3mm}
	\caption{Test accuracy versus efficiency on the ogbn-arxiv dataset.}
	\label{perf_effi}
 	% \vspace{-2mm}
\end{figure}

\para{Performance-efficiency comparison.}
% \zwt{补上ogbn-arxiv数据集上的效率和acc对比图}
In this part, we evaluate the node classification accuracy of NAFS-mean and several baseline methods on the ogbn-arxiv dataset~\citep{hu2020ogb}. 
The predictive accuracy and runtime results are both shown in Figure~\ref{perf_effi} whose X-axis is in log scale.
The Figure shows that NAFS-mean only falls behind GAT and outperforms many competitive baselines in terms of test accuracy. 
Although the test accuracy of GAT is slightly higher than NAFS-mean, NAFS-mean achieves over 6x speedup than GAT.

\subsection{Scalability Comparison on Synthetic Graphs}
\label{scalability_comp}
%\zwt{自己造数据集，随着size增大，慢慢的GCN和GAT不能跑了，GAE和AGE不能跑了，画个Table对错号}
%\zwt{和5，5和一起，efficiency一个图但是不同的时间，scalability就是随着size增大别人不能比，参考Grain}
To test the scalability of our proposed NAFS, we also use the Erdős-Rényi graph generator in the Python package NetworkX~\citep{hagberg2008exploring} to generate artificial graphs of different sizes.
The node sizes of the generated artificial graphs vary from 5,000 to 100,000, and the probability of an edge exists between two nodes is set to 0.0001.
The experiment settings are altered compared to Sec.~\ref{effi_compare}: both GAE and AGE are trained on an NVIDIA TITAN RTX, which has 24 GB of memory since only using CPU to train is unacceptable on large graphs in real-world applications.

The overall experiment results are shown in Figure~\ref{fig.scalability}. 
Figure~\ref{scala_effi} shows a more detailed comparison on relatively small artificial graphs whose node sizes range from 5,000 to 20,000.
Besides, Figure~\ref{scala_scala} shows that GAE encounters the out-of-memory problem on the artificial graph composed of 30,000 nodes, while AGE encounters the out-of-memory problem at graph size 80,000. 
Our proposed NAFS is successfully carried out on all the artificial graphs.
% Note the y-axis in Fig.~\ref{scala_scala} is on the log scale.
On the artificial graph consisting of 100,000 nodes, NAFS accomplishes the node embedding generation in 66.1 seconds, which is less than the running time of AGE on the artificial graph consisting of 10,000 nodes, 80.9 seconds.
The graph size limit of our proposed NAFS is bounded by the CPU memory size.
As long as one can successfully execute the multiplication of the sparse adjacency matrix and the feature matrix, NAFS can then be implemented.
%Moreover, since this multiplication is the core process of message passing, NAFS is at least as scalable as the most scalable one among all the message-passing-based graph embedding methods.

\subsection{Performance on sparse graphs}
% \zwt{加上openreview里在PubMed上补充的关于稀疏图的实验}
To validate the performance benefits of NAFS on sparse graphs, we conduct experiments on the PubMed dataset under two handcrafted sparsity settings: edge sparsity and feature sparsity.

\para{Two sparsity settings.} Under the edge sparsity setting, we randomly remove some edges in the original PubMed dataset to strengthen the edge sparsity issue. 
The edges removed from the original graph are kept the same across all the compared methods under the same edge removing rate.
Under the feature sparsity setting, we randomly choose some nodes in the original PubMed dataset and set their feature vectors to all-zero vectors. 
The selected nodes are kept the same across all the compared methods under the same feature removing rate.

\para{Experiment settings.} We report the performance of GAE and NAFS-mean on the link prediction task under different sparsity settings.
We adopt two metrics - Area Under Curve (AUC) and Average Precision (AP) to evaluate the performance of each method.
The evaluations are conducted with the edge removing rate and the feature removing rate set to 0.2, 0.4, 0.6, respectively. 
We repeat each method 10 times and report the mean performance and the corresponding standard deviations in Table~\ref{table.edge_sparse_unsup} and~\ref{table.feat_sparse_unsup}.

\para{Experiment results.} The experimental results from Table~\ref{table.edge_sparse_unsup} and~\ref{table.feat_sparse_unsup} show that 1$)$ under both edge and feature sparsity settings, NAFS consistently outperforms GAE and GCN on the link prediction task and the node classification task, respectively. 
2$)$ the performance gains are larger in sparser graphs with larger feature/edge removing rates. 
Concretely, the AUC of NAFS outperforms GAE by a margin of 1.0\% if the edge is not removed, and the performance gain has increased to 3.7\% when the edge removing rate is 0.6.

\begin{table}[tpb!]
% \vspace{-2mm}
\centering
{
\caption{Performance comparison under different \emph{edge removing rates} on the PubMed dataset.}
\label{table.edge_sparse_unsup}
% \vspace{2mm}
\renewcommand{\multirowsetup}{\centering}
\resizebox{0.95\linewidth}{!}{
\begin{tabular}{cccccc}
\toprule
\textbf{Methods}&\textbf{Metrics}&\textbf{0.0}& \textbf{0.2}&\textbf{0.4}&
\textbf{0.6}\\
\midrule
\multirowcell{2}{GAE}&
AUC& 96.4$\pm$0.4 & 93.4$\pm$0.6 & 92.6$\pm$0.5 & 90.6$\pm$0.5  \\
&AP& 96.5$\pm$0.5 & 93.7$\pm$0.4 & 92.4$\pm$0.6 & 90.4$\pm$0.5   \\
\midrule
\multirowcell{2}{NAFS-mean}&
AUC& 97.4 (+1.0) & 96.9 (+3.5) & 95.9 (+3.3) & 94.3 (\textbf{+3.7}) \\
&AP& 97.2 (+0.7) & 96.4 (+2.7) & 95.2 (+2.8) & 93.5 (\textbf{+3.1}) \\
\bottomrule
\end{tabular}}}
\end{table}

\begin{table}[tpb!]
% \vspace{-2mm}
\centering
{
\caption{Performance comparison under different \emph{feature removing rates} on the PubMed dataset.}
\label{table.feat_sparse_unsup}
% \vspace{2mm}
\renewcommand{\multirowsetup}{\centering}
\resizebox{0.95\linewidth}{!}{
\begin{tabular}{cccccc}
\toprule
\textbf{Methods}&\textbf{Metrics}&\textbf{0.0}& \textbf{0.2}&\textbf{0.4}&
\textbf{0.6}\\
\midrule
\multirowcell{2}{GAE}&
AUC& 96.4$\pm$0.4 & 89.5$\pm$0.6 & 83.5$\pm$0.5 & 77.4$\pm$0.5  \\
&AP& 96.5$\pm$0.5 & 90.4$\pm$0.4 & 85.5$\pm$0.6 & 80.9$\pm$0.5   \\
\midrule
\multirowcell{2}{NAFS-mean}&
AUC& 97.4 (+1.0) & 93.1 (+3.6) & 87.7 (\textbf{+4.2}) & 81.7 (+3.7) \\
&AP& 97.2 (+0.7) & 94.4 (+4.0) & 91.0 (+5.5) & 86.8 (\textbf{+5.9}) \\
\bottomrule
\end{tabular}}}
\end{table}

The experiment results illustrate that NAFS can effectively exploit distant neighborhood information without the over-smoothing problem and thus get better performance on sparse graphs compared with baseline methods.

\subsection{Comparison with NDLS}
NAFS differs from the related work NDLS in:
1) NDLS searches for the optimal smoothing step for each node, while NAFS presents ``smoothing weight'' to aggregate the outputs after different smoothing steps.
2)  Compared with NDLS, NAFS further proposes to combine the smoothed outputs of different smoothing operators.
3) NDLS only focuses on the semi-supervised node classification task, while NAFS further considers the unsupervised task.
The comparison in Table~\ref{exp} shows NAFS's performance superiority over NDLS on unsupervised tasks.

\begin{table}[t]\caption {Comparison with NDLS on different tasks.}
% \vspace{-1.5em}
% \renewcommand{\thesubtable}{\footnotesize(\alph{subtable})}
\linespread{0.95}
\centering
\setlength{\tabcolsep}{6mm}{
\subtable[Node Clustering]{
% \vspace{-1.8em}
\resizebox{0.36\pdfpagewidth}{!}{
\begin{tabular}{c|c|cccc}
\toprule
\textbf{Methods}&\textbf{Metrics}&\textbf{Cora}& \textbf{Citeseer}&\textbf{PubMed}&\textbf{Wiki}\\
\midrule
\multirowcell{3}{NDLS}&
ACC& 70.6 & 67.5 & \textbf{70.9} & 36.6 \\
&NMI& 52.9 & 41.2 & \textbf{34.6} & 36.1\\
&ARI& 47.4 & 43.5 & \textbf{34.3} & 18.2\\
\midrule
\multirowcell{3}{NAFS-concat}&
ACC& \textbf{75.4} & \textbf{71.1} & 70.5 & \textbf{53.6} \\
&NMI& \textbf{58.6} & \textbf{45.8} & 33.9 & \textbf{50.5} \\
&ARI& \textbf{53.8} & \textbf{46.1} & 33.2 & \textbf{26.3} \\
\bottomrule
\end{tabular}
    }
}
\newline
\vspace{-1.1em}
\newline
}
\setlength{\tabcolsep}{6mm}{
\subtable[Link Prediction]
{
\vspace{-1.8em}
\resizebox{0.36\pdfpagewidth}{!}{
\vspace{-1.8em}
\begin{tabular}{c|c|ccc}
\toprule
\textbf{Methods}&\textbf{Metrics}&\textbf{Cora}& \textbf{Citeseer}&\textbf{PubMed}\\
\midrule
\multirowcell{2}{NDLS}&
AUC& 90.6 & 92.5 & 95.3 \\
&AP& 91.3 & 92.8 & 95.1 \\
\midrule
\multirowcell{2}{NAFS-concat}&
AUC& \textbf{92.6} & \textbf{93.7} & \textbf{97.6} \\
&AP& \textbf{93.8} & \textbf{93.1} & \textbf{97.2} \\
\bottomrule
\end{tabular}
    }
}
}
\vspace{-2.7em}
\label{exp}
\end{table}

\section{More details of NAFS}
\label{details}
\subsection{Pseudo code of NAFS}
\label{alg}
\begin{algorithm}[t]
  \SetAlgoLined
  \KwIn{Smoothing step $K$, feature matrix $\mathbf{X}$, adjacency matrix $\mathbf{A}$, and $\{r_1, r_2,...,r_T\}$.}
  \KwOut{Graph embedding matrix $\mathbf{Z}$.}
  \SetAlgoLined
  \caption{\textbf{\sys pipeline.}}
  \label{alg:graphfe}
   Initialize the feature matrix $\mathbf{X}^{(0)} = \mathbf{X}$;
   
   \textbf{Operation 1: Feature Smoothing}
   
   \For{$1\leq t\leq T$}
   {
       Update the normalized adjacency matrix $\mathbf{\hat{A}}_{r_t}=
       \widetilde{\mathbf{D}}^{r_t-1}\widetilde{\mathbf{A}}\widetilde{\mathbf{D}}^{-r_t}$ ;
       
       \For{$1\leq i\leq n$}
       {
            \For{$0\leq k\leq K$}
            {
               Calculate $D_{i}(k)$ and $w_{i}(k)$ with Eq.~\ref{dik} and~\ref{iw}, respectively;
            }
       }
       \For{$0\leq k\leq K$}
       {
            Construct $\mathbf{W}(k)$ with Eq.~\ref{w_k};
       }
       Smooth the node features $\mathbf{X}$ with $\hat{\mathbf{X}}^{(t)} = \sum\limits_{k=0}\limits^{K}\mathbf{W}(k)\hat{\mathbf{A}}_{r_t}^{k}\mathbf{X}$;  
   }

    \textbf{Operation 2: Feature Ensemble}
    
    Compute the final embedding $\mathbf{Z} $ with $\mathbf{Z}  \gets \oplus_{i\in \{1, 2, ..., T\} }\hat{\mathbf{X}}^{(i)}$. 
\end{algorithm}
% Alg.~\ref{alg:graphfe} presents an overview of the proposed method.
% \zwt{则昂补充对伪代码的描述}
Alg.~\ref{alg:graphfe} shows the whole pipeline of our proposed \sys. 
We first initialize $\mathbf{X}^{(0)}$ as the original feature matrix $\mathbf{X}$. 
Given the normalization parameter $r_t$, we obtained the corresponding normalized adjacency matrix $\mathbf{\hat{A}}_{r_t} = \widetilde{\mathbf{D}}^{r_t-1}\widetilde{\mathbf{A}}\widetilde{\mathbf{D}}^{-r_t}$, which acts as knowledge extractors (line 4). 
After that, for each node, we use E.q.~\ref{dik} to calculate the Over-smoothing Distance with all the $k$ ranging from $0$ to $K$ (line 6, 7). 
Then we calculate its Aggregation Weights through Eq.~\ref{iw} (line 8, 9). 
After obtaining Aggregation Weights with all $k$ and $i$, we construct the Aggregation Weight matrix for each $k$ with Eq.~\ref{w_k} (line 10, 11). 
Next, we compute the NAFS output $\hat{\mathbf{X}}^{(t)}$ with $\hat{\mathbf{X}}^{(t)} = \sum\limits_{k=0}\limits^{K}\mathbf{W}(k)\hat{\mathbf{A}}_{r_t}^{k}\mathbf{X}$ (line 12). 
Finally, we compute the final embedding result of all $t$ through $\mathbf{Z}  \gets \oplus_{i\in \{1, 2, ..., T\} }\hat{\mathbf{X}}^{(i)}$ (line 14).

\subsection{Motivation of Feature Ensemble} 
Adopting different smoothing operators in the feature smoothing operation (the normalized adjacency matrix $\mathbf{\hat{A}}$ in Eq.~\ref{eq_GC}) is equivalent to smoothing features in different manners.
Other than the normalized adjacency matrix $\mathbf{\hat{A}}$, there are many alternatives that have been proposed recently. For example,
GraphSAGE~\citep{hamilton2017inductive} designs three smoothing operators (i.e., Mean, LSTM and Pooling) to flexibly capture the information of neighboring nodes.
SIGN~\citep{frasca2020sign} enriches the smoothing operators with Personalized-PageRank-based~\citep{klicpera2019diffusion} and triangle-based~\citep{monti2018motifnet} adjacency matrices.
However, these methods are not designed for graph representation learning, and different smoothing operators may result in diverse smoothed features.
For better node representation, the feature ensemble is used to combine the smoothed feature under different smoothing operators.

\subsection{Relations with other Scalable GNN Architectures}
The scalable GNNs can be roughly classified into two categories: (a) "first propagate then predict"; (b) sampling + ordinary GNN.
The first category includes famous GNN models like SGC~\citep{wu2019simplifying} and SIGN~\citep{frasca2020sign}. 
They disentangle the coupled propagation and transformation operations in traditional GCN layers~\citep{kipf2016semi}, and execute all the propagation operations as preprocessing.
The most representative GNN model of the second category is GraphSAGE~\citep{hamilton2017inductive}, and many other works have been proposed following it, like FastGCN~\citep{chen2018fastgcn} and GraphSAINT~\citep{zeng2019graphsaint}.
The main contribution of these works is effective sampling strategies that can preserve the most valuable information from the original graph.
The sampling strategies always act as a plug-and-play module, and can be combined with ordinary GNNs, which allows the latter to perform on large-scale graphs.
GNNs belong to the "first propagate then predict" category usually enjoy higher efficiency since they avoid recursively performing propagation in each training epoch.
Our proposed NAFS belongs to the "first propagate then predict" category.

\subsection{Advantage in Distributed Settings}
% \red{maybe add an experiment}
NAFS can also be adapted to the distributed environment.
The feature smoothing process in \sys is sparse matrix dense matrix multiplications, which have mature implementations in distributed environments. Besides, this process only needs to be pre-computed at once.
In contrast, during each training iteration of GAE and its variants, each node must repeatedly pull the intermediate representations of other nodes, leading to high communication costs.

\subsection{Association with Mixing Time in Markov Chain}
Setting $r=0$ in the adjacency matrix makes the feature smoothing process a Markov chain.
To study the mixing time at the node level, we denote $t_{mix}(\epsilon, i)$ as the minimum step for node $i$ such that the distance between its representation and its stationary state is at most $\epsilon$.
It can be calculated as follows with the help of Lemma 3 in Appendix A.1:
\vspace{-8pt}
\begin{equation}
\small
\vspace{-5mm}
\scriptsize
\begin{split}
t_{mix}(\epsilon, i)
&= \min_{t \in \mathbb{N}} \left\{d(t, i) \leq \epsilon \right\}\\
&= \min_{t \in \mathbb{N}} \left\{|P^t(i,\cdot)-\pi ||_{TV} \leq \epsilon \right\} \\
&\leq \min_{t \in \mathbb{N}} \left\{\frac{\lambda_2^t\sum_{j=1}^n \widetilde{d_j}|\mathbf{X}_j|}{2\widetilde{d_i}} \leq \epsilon \right\}
= \lceil \log_{\lambda_2} \frac{2\widetilde{d_i}\epsilon}{\sum_{j=1}^n \widetilde{d_j}|\mathbf{X}_j|}\rceil.\\
\end{split} \nonumber
\end{equation}
Intuitively, $t_{mix}(\epsilon, i)$ is negatively correlated with ``smoothing speed''.
And the upper bound shows that nodes with smaller degrees possess larger $t_{mix}(\epsilon, i)$s, which verifies the above assumption: the nodes with smaller degrees own slower "smoothing speeds" as shown in Figure~\ref{smoothing_speed}.

\section{Reproduction}
The source code of \sys can be found in Anonymous Github (\url{https://github.com/PKU-DAIR/NAFS}). 
To ensure reproducibility, we have provided the overview of datasets and baselines in Sec.~\ref{sec:datasets} and Appendix B.1. 
The settings for each task and baseline can be found in Sec.~\ref{sec:setups}.
Our experimental environment is presented in Appendix B.2, and please refer to ``README.md'' in the Github repository for more reproduction details.

\end{document}